\documentclass{article}
\usepackage[nonatbib,preprint]{nips_2018}

\usepackage[utf8]{inputenc} 
\usepackage[T1]{fontenc}    
\usepackage{hyperref}       
\usepackage{url}            
\usepackage{booktabs}       
\usepackage{amsfonts}       
\usepackage{nicefrac}       
\usepackage{microtype}      

\usepackage{graphicx}
\graphicspath{ {images/} }

\def\vect#1{\mbox{\boldmath $#1$}}

\newcommand{\argmin}{\mathop{\rm arg~min}\limits}

\title{Disentangling Controllable and Uncontrollable Factors of Variation by Interacting with the World}

\author{
  Yoshihide Sawada\\
  Panasonic Corporation\\
  \texttt{sawada.yoshihide@jp.panasonic.com} \\
}

\begin{document}

\maketitle

\begin{abstract}
We introduce a method to disentangle controllable and uncontrollable factors of variation by interacting with the world.
Disentanglement leads to good representations and is important when applying deep neural networks~(DNNs) in fields where explanations are required.
This study attempts to improve an existing reinforcement learning~(RL) approach to disentangle controllable and uncontrollable factors of variation, because the method lacks a mechanism to represent uncontrollable obstacles.
To address this problem, we train two DNNs simultaneously: one that represents the controllable object and another that represents uncontrollable obstacles.
For stable training, we applied a pretraining approach using a model robust against uncontrollable obstacles.
Simulation experiments demonstrate that the proposed model can disentangle independently controllable and uncontrollable factors without annotated data.
\end{abstract}

\section{Introduction}
Deep learning is a family of machine learning method used in a wide variety of applications, such as medical assistance, self-driving cars, and natural science. 
Although, deep neural networks~(DNNs) have achieved state-of-the-art performances, and have produced results superior to human expert' perfection, however, their results are difficult for humans to explain due to entangled nature.
Interpretability is important when applying DNNs in the fields where explanations are required.
In this study, we investigate a method to improve this issue.

Interpretability is defined as the ability to explain or provide meaning in human-understandable~\cite{46160,guidotti2018survey}.
Interpretability can be realized by disentangling meaningful factors of variation, which means that each neurons is sensitive to changes in its corresponding factor of variation and relatively invariant to changes in the others~~\cite{reed2014learning}.
Many studies have focused on disentangled generative models based on a generative adversarial network (GAN)~\cite{li2017infogail,mathieu2016disentangling,tran2017disentangled} and a variational autoencoder (VAE)~\cite{chen2018isolating,dupont2018joint,higgins2016beta}.
These methods seem to work well, however, it is still difficult to always give independent and meaningful information to neurons without labor-intensive manual annotation.

In the present study, we employ the reinforcement learning~(RL) based method proposed by Thomas et al.~\cite{thomas2017independently,thomas2017independently2}.
This method was inspired by humans learning through an interaction, which did not require extrinsic rewards.
Thomas et al. trained their model by constraining the highest hidden layer in which each neuron reacted independently with one of the controllable object's actions.
Although this method does not require annotated training data, it still lacks an important mechanism in generating good representations, i.e., it cannot disentangle the controllable and uncontrollable factors of variation.
Disentangling these factors is also important in many application fields (e.g., self-driving cars).
Thus, we propose a method to address this issue.

\begin{figure}[t]
\centering
\includegraphics[width=0.65\linewidth]{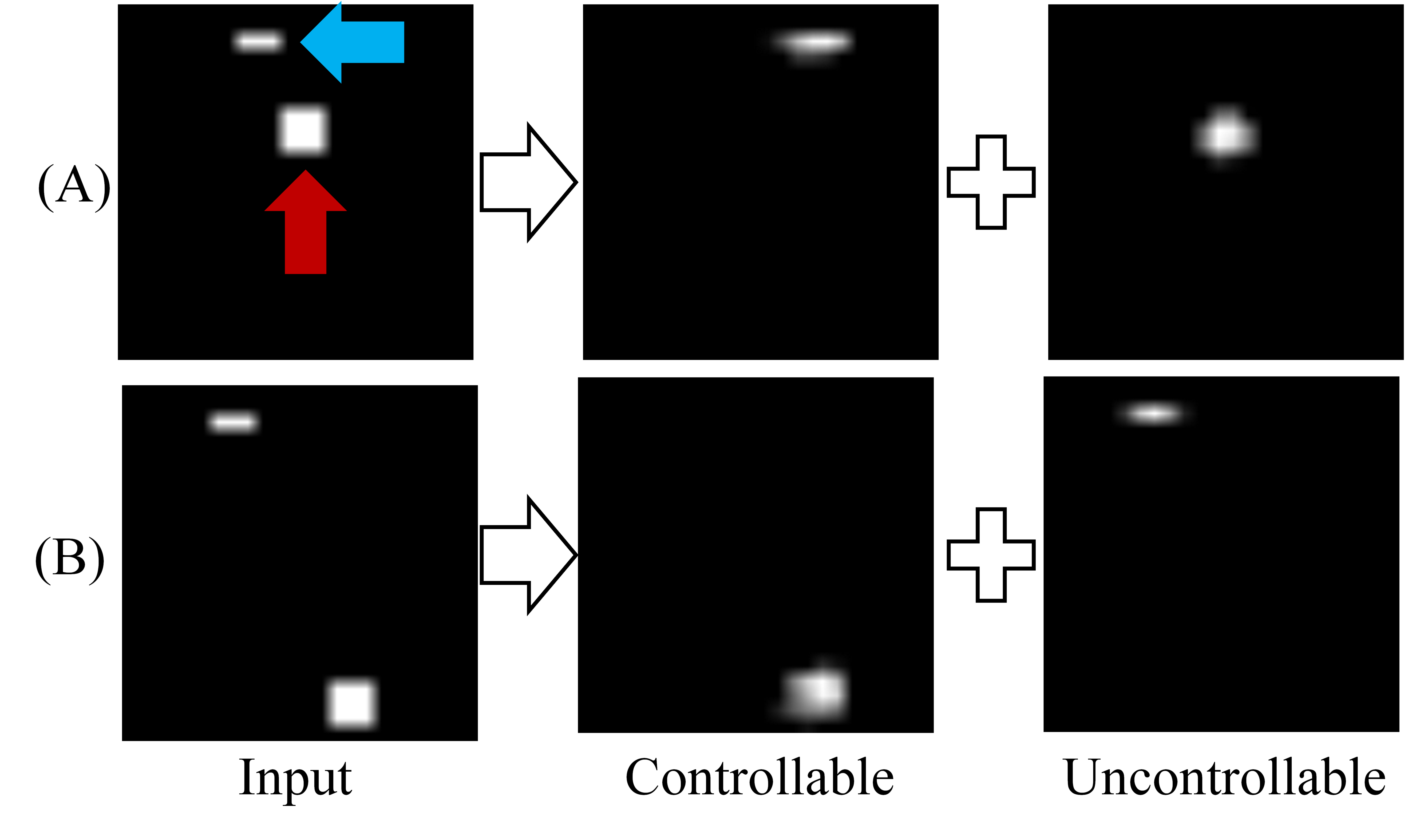}
\caption{Examples of reconstruction results. Red and blue arrows represent controllable and uncontrollable objects, respectively. (A) A model that incorrectly identifies the two objects. (B) The inconsistency is resolved by pretraining.}
\label{fig:reconst_fail_ex}
\end{figure}

The proposed method involves training two DNNs simultaneously, where one network represents the controllable object based on the Thomas' model, and the second method represents uncontrollable obstacles.
Similar to the Thomas' model, we adopt a strategy that does not require manual annotation.
However, we found that this strategy makes stable training difficult.
Namely, trained two DNNs were unable to accurately differentiate between controllable and uncontrollable objects~(Fig. \ref{fig:reconst_fail_ex}~(A) shows an example).
To address this issue, we focus on one of the good aspects of the Thomas' model, i.e., robustness against uncontrollable obstacles.
Our analysis indicates that the Thomas' method does not provide an incentive to encode environmental features unrelated to the controllable object's actions.
From this, we found that trained DNN could ignore uncontrollable obstacles even though it did not have a disentanglement mechanism for both controllable and uncontrollable objects.
Based on this robustness, we use this mechanism to set the initial DNN parameters for the controllable object to constrain neurons to react to the controllable object's actions.
Using this pretraining approach, we confirm that the proposed model can disentangle the controllable and uncontrollable factors of variation without annotation.

We use simulations to evaluate the proposed model, and the results demonstrate that the model can disentangle the controllable and uncontrollable factors of variation.
In addition, we investigate the behaviour of the proposed model on a task that involves an acquisition of extrinsic rewards.


The study provides the following major contributions:
\begin{itemize}
\item We analyze the Thomas' model in an environment including uncontrollable obstacles.
\item We propose a disentangled model constructed with a DNN that represents the controllable object based on the Thomas' model and a DNN that represents the uncontrollable obstacles.
\item For stable training, we propose reusing the trained Thomas' model as an initial parameter of the DNN of the controllable object.
\end{itemize}

\section{Related Work}
Interpretability is a hot topic in the deep learning field, and research has focused on two primary approaches: finding ways to understand a black-box DNNs and learning disentangled representation.

Understanding black-box DNNs is an attempt to explain trained layers that perform well but lack interpretability~\cite{guidotti2018survey,2018arXiv180200614Z}.
For example, Alain and Bengio~\cite{alain2016understanding} used linear probes to analyze information dynamics, and Bau et al.~\cite{netdissect2017} used feature maps to visualize the input image regions informative for classification.
In addition, Shrikumar et al.~\cite{shrikumar2017learning} computed importance scores based on reference inputs and outputs, and Koh and Liang~\cite{koh2017understanding} proposed approximated influence functions to understand the influence of any training data on the model.
While these approaches can yield high performance, they attempt to explain even uninterpretable features caused by an entangled DNN.

Learning disentangled representations aim to construct a small number of independent and meaningful features of the input data to deliver good representations. 
This has been attempted with (semi)-supervised, unsupervised, and RL approaches.

All supervised learning-based approaches use the labeled information to disentangle features.
For example, Reed et al.~\cite{reed2014learning} proposed a restricted Boltzmann machine-based method that considered a complicated manifolds as collections of sub-manifolds. 
Narayanaswamy et al.~\cite{NIPS2017_7174} combined a semi-supervised VAE~\cite{kingma2013auto} and a probabilistic graphical model.
Peng et al.~\cite{peng2017reconstruction} proposed a hierarchical model that first disentangled identity and non-identity features from face images and then disentangled poses and landmarks from the non-identity features.
Tran et al.~\cite{tran2017disentangled} proposed a disentangled representation learning method for pose-invariant face recognition based on a GAN \cite{goodfellow2014generative}.
To disentangle the GAN generator's input variables, Mathieu et al.~\cite{mathieu2016disentangling} used a VAE, and Spurr et al.~\cite{spurr2017guiding} proposed a semi-supervised InfoGAN.

The most studied unsupervised learning-based methods employ generative models, especially VAEs and InfoGANs~\cite{chen2016infogan}.
VAEs combine Bayes and autoencoder approaches to embed encoded features based on a given probability distribution.
Dupont~\cite{dupont2018joint} proposed a joint-VAE that disentangled continuous and discrete representations. 
Higgins et al.~\cite{higgins2016beta} introduced an adjustable hyperparameter, $\beta > 1$, to a standard VAE to balance latent channel capacity and independence constraints with reconstruction accuracy.
Kim and Mnih~\cite{kim2018disentangling} added a discriminator to $\beta$-VAE to estimate Total Correlation, and Chen et al.~\cite{chen2018isolating} decomposed a $\beta$-VAE equation and refined it to improve the disentanglement ability without using additional hyperparameters. 
InfoGANs attempted to learn a generator such that they cheated the discriminator and maximized the mutual information between synthetic samples and newly introduced latent codes.
Inspired by InfoGAN, Li et al.~\cite{li2017infogail} proposed combining generative adversarial imitation learning~\cite{ho2016generative} with a variational lower bound of the mutual information.
On the application side, for example, unsupervised approaches for sequential data~\cite{denton2017unsupervised,fraccaro2017disentangled,hsu2017unsupervised}, and control and planing~\cite{banijamali2017disentangling,li2017infogail} were also proposed.

As the RL-based approaches, Mao et al.~\cite{mao2018universal} proposed a method for effective transfer learning by disentangling task-specific and environment-specific information using start and goal states.
Meanwhile, we focus on learning good representations which can disentangle meaningful factors of variation included in the environment.
Namely, we train the model that can disentangle controllable and uncontrollable factors of variation while disentangling independently controllable factors.

\section{Method}

This study focuses on the model proposed by Thomas et al.~\cite{thomas2017independently,thomas2017independently2}, which was inspired by humans learning through interaction.
They proposed two approaches for disentangle learning, i.e., one that corresponds to each neuron to each action~\cite{thomas2017independently2} and another that corresponds each vector generated from the latent features to each action~\cite{thomas2017independently,thomas2017independently2}.
To attribute independent meaning to each neuron, we focus on the approach that corresponds to each neuron to each action.

\subsection{Disentangling Independently Controllable Factors of Variation}
\label{sub:thomas_equ}
Thomas' method constrains each neuron in the highest hidden layer to react to one of the controllable object's actions independently while reconstructing the controllable object.
\begin{equation}
{\cal L}(\vect{\phi}, \vect{\theta}, \vect{\Psi}) = \sum_i {\cal R}_i( \vect{\phi}, \vect{\theta}) - \lambda \sum_k {\cal S}_{i,k}( \vect{\phi}, \vect{\Psi}), 
\label{equ:icfv}
\end{equation}
where $\vect{\phi}$ and $\vect{\theta}$ represent the parameters of encoder $f$ and decoder $g$. 
${\cal R}_i$ represents the reconstruction error for the $i$-th observed data point $\vect{x}_i$ in the current state $s_i$:
\begin{equation}
{\cal R}_i(\vect{\phi}, \vect{\theta}) = \| \vect{x}_i - g(f(\vect{x}_i)) \|^2.
\label{equ:reconst}
\end{equation}
In Eq.~(\ref{equ:icfv}), ${\cal S}_i$ is a weighted selectivity function.
Selectivity function $sel_{i,k}$ is related to the intrinsic rewards proposed by Pathak et al.~\cite{pathak2017curiosity}.
Pathak et al. defined intrinsic rewards as $\| f(\vect{x}') - f(\vect{x}) \|^2$ while $sel_{i,k}$ was element-wise normalization to disentangle independently controllable factors of variation.
\begin{eqnarray}
{\cal S}_{i,k}( \vect{\phi}, \vect{\Psi} ) &=& 
\sum_{a} \pi_{\psi_k}(a \mid f(\vect{x}_i)) sel_{i,k} (\vect{\phi}, a) \nonumber \\
&=& \sum_{a} \pi_{\psi_k}(a \mid f(\vect{x}_i)) \sum_{s'} \left[ \log \left( \frac{1}{K} +  \frac{\mid f_{k}(\vect{x}')-f_{k}(\vect{x}_i) \mid}{\sum_{k'} \mid f_{k'}(\vect{x}')-f_{k'}(\vect{x}_i) \mid} \right) \right],
\label{lab:selectivity_function}
\end{eqnarray}
where $\vect{x}'$ represents the observed data point in the next state $s' \sim P_{s,s'}^{\pi_{\psi_k}}$, and $P_{s,s'}^{\pi_{\psi_k}}$ represents the environment transition distribution from $s$ to $s'$ under $\pi_{\psi_k}(a \mid f(\vect{x}))$ parameterized by the $k$-th parameter $\vect{\psi}_k$. 
$\vect{\Psi} = \{ \vect{\psi}_k \mid k = 1, 2, \cdots, K \}$ where $K$ is the number of actions, $f(\vect{x}) \in \mathbb{R}^K$, and $\pi_{\psi_k}(a \mid f(\vect{x})) \in \mathbb{R}^K$ is the policy~(weight) to support that the $k$-th output neuron $f_k(\vect{x})$ becomes the highest value under action $a$.
Note that many variations of ${\cal S}_{i,k}$ are possible~\cite{thomas2017independently2}.
We derived Eq.~(\ref{lab:selectivity_function}) from a equation which links the lower bound of the mutual information~\cite{thomas2017independently}, which is discussed in the Supplementary Materials.

Adding ${\cal S}_{i,k}$ to the reconstruction error, it means that the $k$-th neuron in the trained encoder $f$ reacts to only action $a$.
Considering that these actions can be interpreted by humans, this model can produce independently interpretable neurons $f_k(\vect{x})$.

\subsection{Disentangling Controllable and Uncontrollable Factors of Variation}
\label{sub:proposed}

\begin{figure}[t]
\centering
\includegraphics[width=0.65\linewidth]{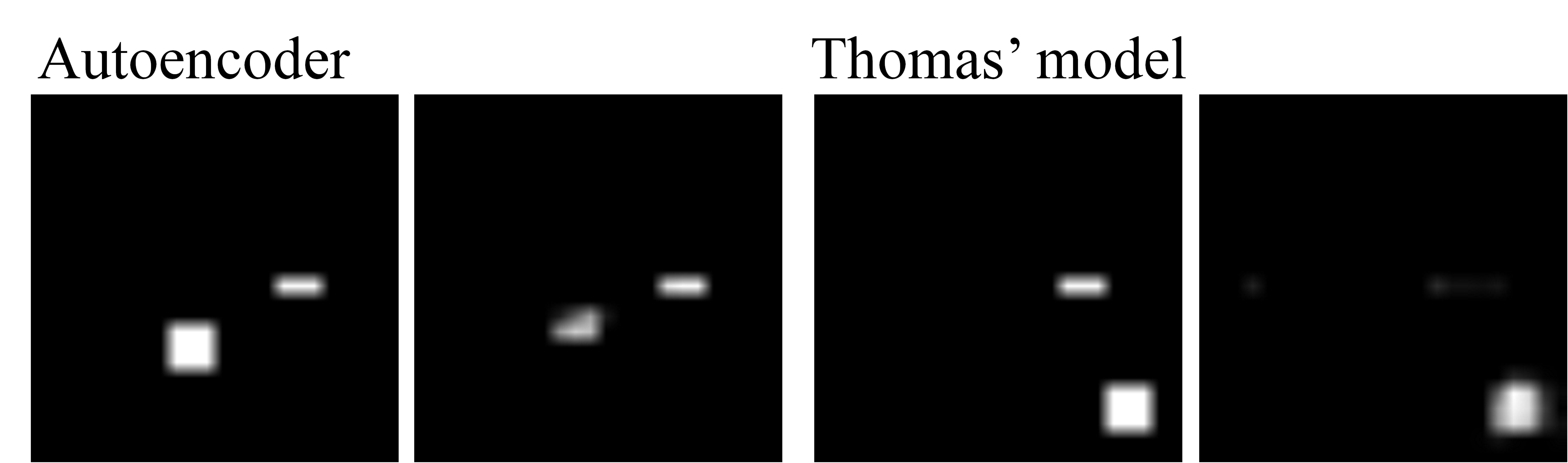}
\caption{Example of reconstruction results of the autoencoder and the Thomas' model. The autoencoder attempted to reconstruct both objects, whereas the Thomas' model only reconstructed the controllable object.}
\label{fig:result_ae_icfv}
\end{figure}

\begin{figure}[t]
\centering
\includegraphics[width=0.65\linewidth]{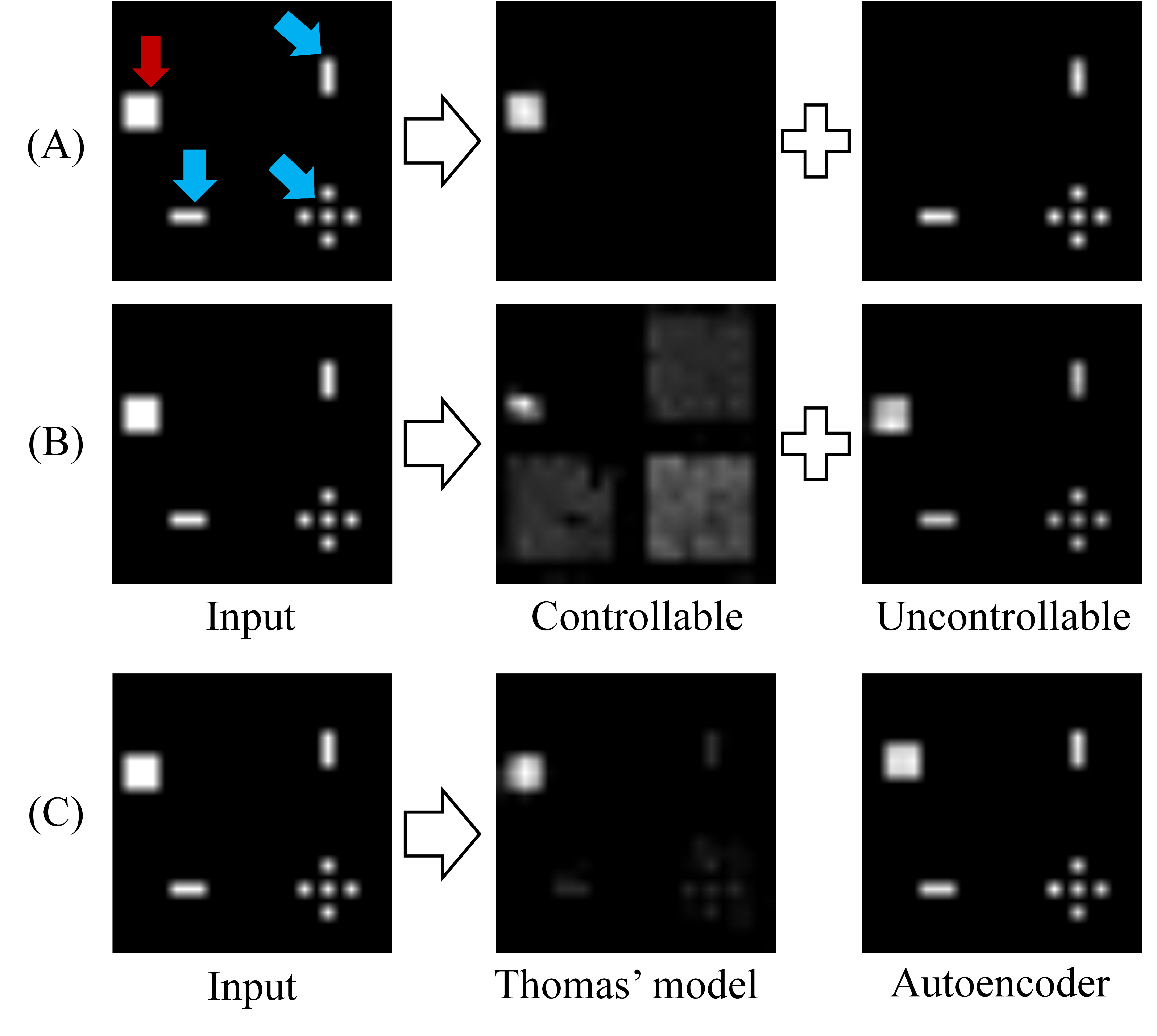}
\caption{Example of reconstruction results for situation 2: (A) proposed model with pretraining, (B) model without pretraining, and (C) autoencoder and the Thomas' model.}
\label{fig:reconst_4_objects}
\end{figure}

In this study, we consider an example in which this method is applied to an environment that contains obstacles.
From equations in Section~\ref{sub:thomas_equ}, the Thomas' method does not have a mechanism for disentangling controllable and uncontrollable factors of variation.
Thus, to disentangle controllable and uncontrollable factors of variation, the proposed method trains two DNNs simultaneously~(one represents the controllable object and another represents uncontrollable obstacles).
\begin{eqnarray}
{\cal L}(\vect{\Phi}, \vect{\Theta}, \vect{\Psi}) = \sum_i \hat{{\cal  R}}_i( \vect{\Phi}, \vect{\Theta}) - \lambda \sum_k {\cal S}_{i,k}( \vect{\phi}_c, \vect{\Psi}), \\
\hat{{\cal R}}_i(\vect{\Phi}, \vect{\Theta}) = \| \vect{x}_i - [ g_c(f_c(\vect{x}_i)) + g_u(f_u(\vect{x}_i)) ]\|^2,
\label{equ:proposed}
\end{eqnarray}
where $\Phi = \{ \vect{\phi}_c, \vect{\phi}_u \}$, $\Theta = \{ \vect{\theta}_c, \vect{\theta}_u \}$, and subscripts $c$ and $u$ represent the parameters for the controllable and uncontrollable factors of variation, respectively.

Initially, the above equations appear to work well; however, uncertainty relative to which objects are controllable or uncontrollable remains.
For example, in Fig.~\ref{fig:reconst_fail_ex}~(A), the model did not identify the two objects correctly. 
Note that we do not use labeled data to determine whether an object is controllable or uncontrollable; therefore, the DNNs cannot differentiate objects during training.

To prevent this training instability, we use a pretraining approach that reuses the results of the Thomas' model~(i.e., $\vect{\phi}, \vect{\theta}$, and $\vect{\psi}$) as our initial parameters (i.e., $\vect{\phi}_c, \vect{\theta}_c$, and $\vect{\psi}_c$).
As described in Section~\ref{sub:thomas_equ}, the $k$-th neuron $f_{k}(\vect{x})$ trained by the Thomas' method reacts to one specific action of the controllable object, i.e., these neurons represent coordinates of the controllable object.
Considering a set of input data in which the controllable object simply moves from left to right across the image with $x$-coordinate $x_i$ in the $i$-th image, the image of this object is given by $\vect{x}(x_i)$.
In this case, action $a$ is always {\sf right} and $x_1 < x_2 < \cdots$. 
If the $k$-th neuron reacts to action $a$ from Eq.~(\ref{lab:selectivity_function}), we find that $f_{k}(\vect{x}(x_{i})) < f_{k}(\vect{x}(x_{i+1})) < \cdots$.
This is strong constraint for the latent features and this constraint clearly does not provide incentive to encode environmental features unrelated to the controllable object's actions.
Namely, the encoder is trained such that $f(\vect{x})$ has only information of controllable object's coordinates and the decoder is trained to reconstruct the image from $f(\vect{x})$ without information of obstacles.
Note that this is clearly a difficult task for the decoder.
As a result, it is likely to reach a local minimum that removes the obstacle.
Therefore, the Thomas' model can ignore information about obstacles, which means that it is robust against obstacles.
For this reason, with our pretraining approach, $\vect{\phi}_c, \vect{\theta}_c$, and $\vect{\psi}_c$ can focus only on the controllable object.
Note that we could have used other approaches, e.g., it may be possible to reuse parameters $\vect{\phi}, \vect{\theta}$, and $\vect{\psi}$ from other environments, which is discussed in the Supplementary Materials. 

\section{Experimental Results}
\label{sub:toy_disentangle}

We evaluated the proposed model's ability to disentangle the controllable and uncontrollable factors of variation using toy experiments to accurately grasp our model's ability.
In addition, we investigated the behavior of the proposed model on an RL task with extrinsic rewards.
Here the objective in this article was to analyze the proposed model's behavior rather than achieve state-of-the-art performances.

There have been various works to quantify disentanglement~\cite{kim2018disentangling}.
In this study, we evaluated it using simple approaches, i.e., correlation between $f_c$ and controllable object coordinates~\cite{thomas2017independently2}, and distance and a concentration matrix in the latent space $f_u$.

We compared the proposed model to each part~(the autoencoder which minimizes $\sum \| \vect{x} - g(f(\vect{x})) \|^2$ and the Thomas' model), and a model without pretraining in two environments, i.e., {\it situation} 1~(Figs.~\ref{fig:reconst_fail_ex} and \ref{fig:result_ae_icfv}) and {\it situation} 2~(Fig.~\ref{fig:reconst_4_objects}).
Here, we move the controllable object using $K = 4$ actions (i.e., {\sf left, right, up}, and {\sf down}), and the uncontrollable objects randomly take the same actions.
Note that the environment conditions and network architectures are described in the Supplementary Materials.


\subsection{Comparison of Reconstruction Results}

\begin{table}[t]
\begin{center}
\caption{Correlation coefficients between $f_{c,k}$ and the controllable object's coordinates for the proposed method~(situation 1). Values in parentheses show the difference between the results of the Thomas' model and the proposed model.}
\begin{tabular}{|l|c|c|c|c|}
\hline
Axis & $f_{c, 1}$ & $f_{c, 2}$ & $f_{c, 3}$ & $f_{c, 4}$ \\
\hline \hline
$x$ & $0.892 (-0.005)$ & $-0.838 (0.002)$ & $-0.029 (-0.002)$ & $0.133 (-0.003)$\\
\hline
$y$ & $0.025 (-0.005)$ & $-0.042 (0.0)$ & $-0.853 (0.005)$ & $0.880 (-0.005)$\\
\hline
\end{tabular}
\label{tbl:correlation}
\end{center}
\end{table}

\begin{table}[t]
\begin{center}
\caption{Correlation coefficients between $f_{c,k}$ and the controllable object's coordinates~(situation 2).}
\begin{tabular}{|l|c|c|c|c|}
\hline
Axis & $f_{c, 1}$ & $f_{c, 2}$ & $f_{c, 3}$ & $f_{c, 4}$ \\
\hline \hline
$x$ & $-0.066 (0.0)$ & $0.031 (0.014)$ & $-0.991 (0.029)$ & $0.978 (-0.014)$\\
\hline
$y$ & $-0.983 (0.009)$ & $0.970 (0.003)$ & $-0.042 (-0.002)$ & $0.058 (-0.007)$\\
\hline
\end{tabular}
\label{tbl:correlation_4}
\end{center}
\end{table}

\begin{table}[t]
\begin{center}
\caption{Correlation coefficients between $f_{c,k}$ and the controllable object's coordinates when applying the model without pretraining. Value to the left of the slash is in situation 1 and value to the right is situation 2.}
\begin{tabular}{|l|c|c|c|c|}
\hline
Axis & $f_{c, 1}$ & $f_{c, 2}$ & $f_{c, 3}$ & $f_{c, 4}$ \\
\hline \hline
$x$ & $0.854 / -0.056$ & $0.855 / 0.996$ & $0.068 / -0.997$ & $-0.746 / 0.051$\\
\hline
$y$ & $0.314 / -0.996$ & $0.313 / 0.076$ & $-0.260 / -0.075$ & $0.481 / 0.996$\\
\hline
\end{tabular}
\label{tbl:correlation_without}
\end{center}
\end{table}

Figures~\ref{fig:reconst_fail_ex}, \ref{fig:result_ae_icfv}, and \ref{fig:reconst_4_objects} showed an example of reconstruction results.
Figure~\ref{fig:result_ae_icfv} showed that the autoencoder reconstructed both objects, whereas the Thomas' model only reconstructed the controllable object.
Figure~\ref{fig:reconst_fail_ex}~(A) and (B) show an example of reconstructions obtained from the proposed model without pretraining and with pretraining, respectively.
As can be seen, the Thomas' model effectively resolves the reconstruction ambiguity.
Figure~\ref{fig:reconst_4_objects} also shows the same results in situation 1.
The results demonstrate that our method can consistently controllable and uncontrollable factors of variation.

\subsection{Analysis of DNN for Controllable Object}
\label{sub:controllable}

\begin{figure}[t]
\centering
\begin{minipage}{0.3\hsize}
\begin{center}
\includegraphics[width=1.0\linewidth]{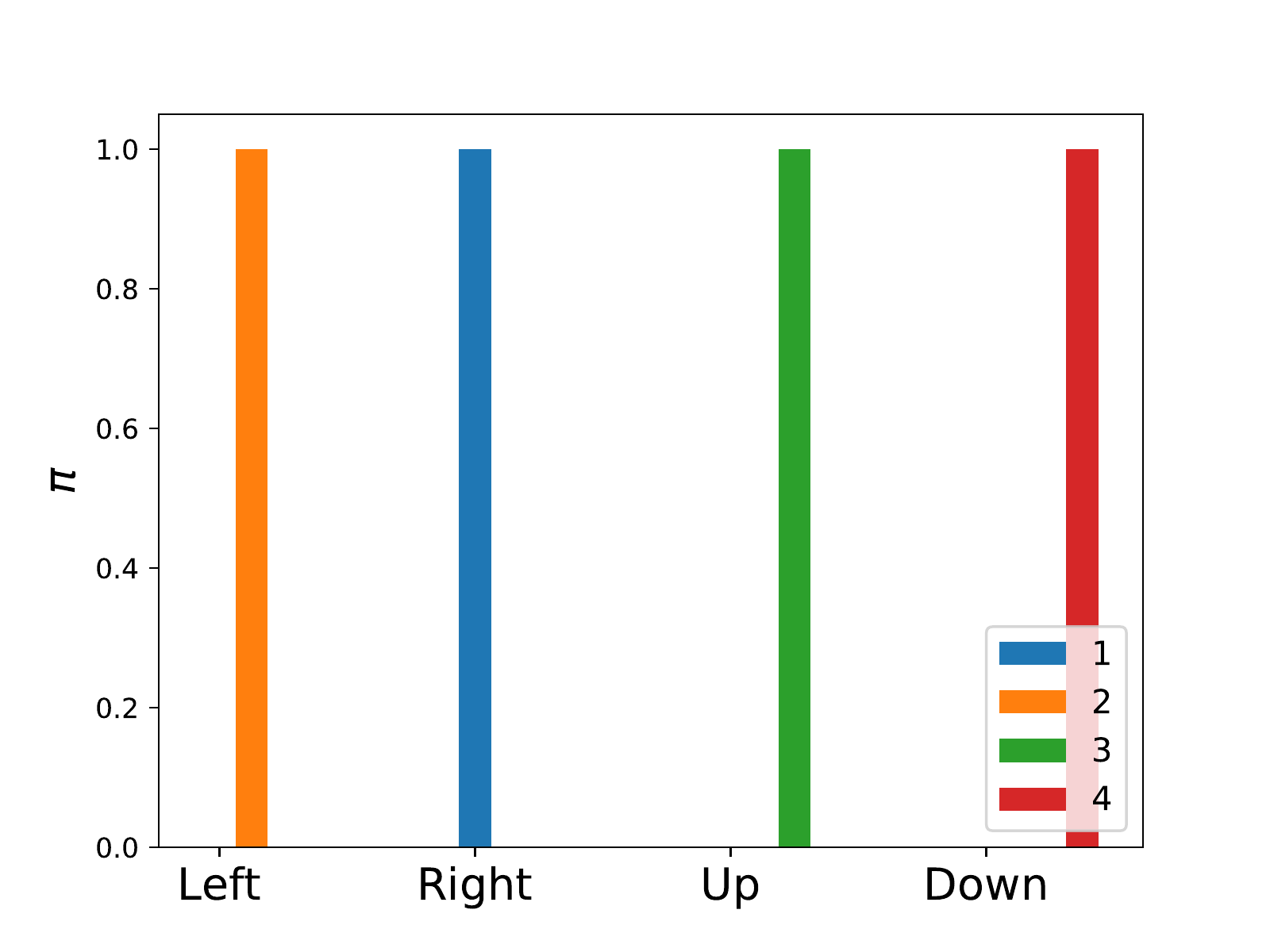}\\
(A)
\end{center}
\end{minipage}
\begin{minipage}{0.3\hsize}
\begin{center}
\includegraphics[width=1.0\linewidth]{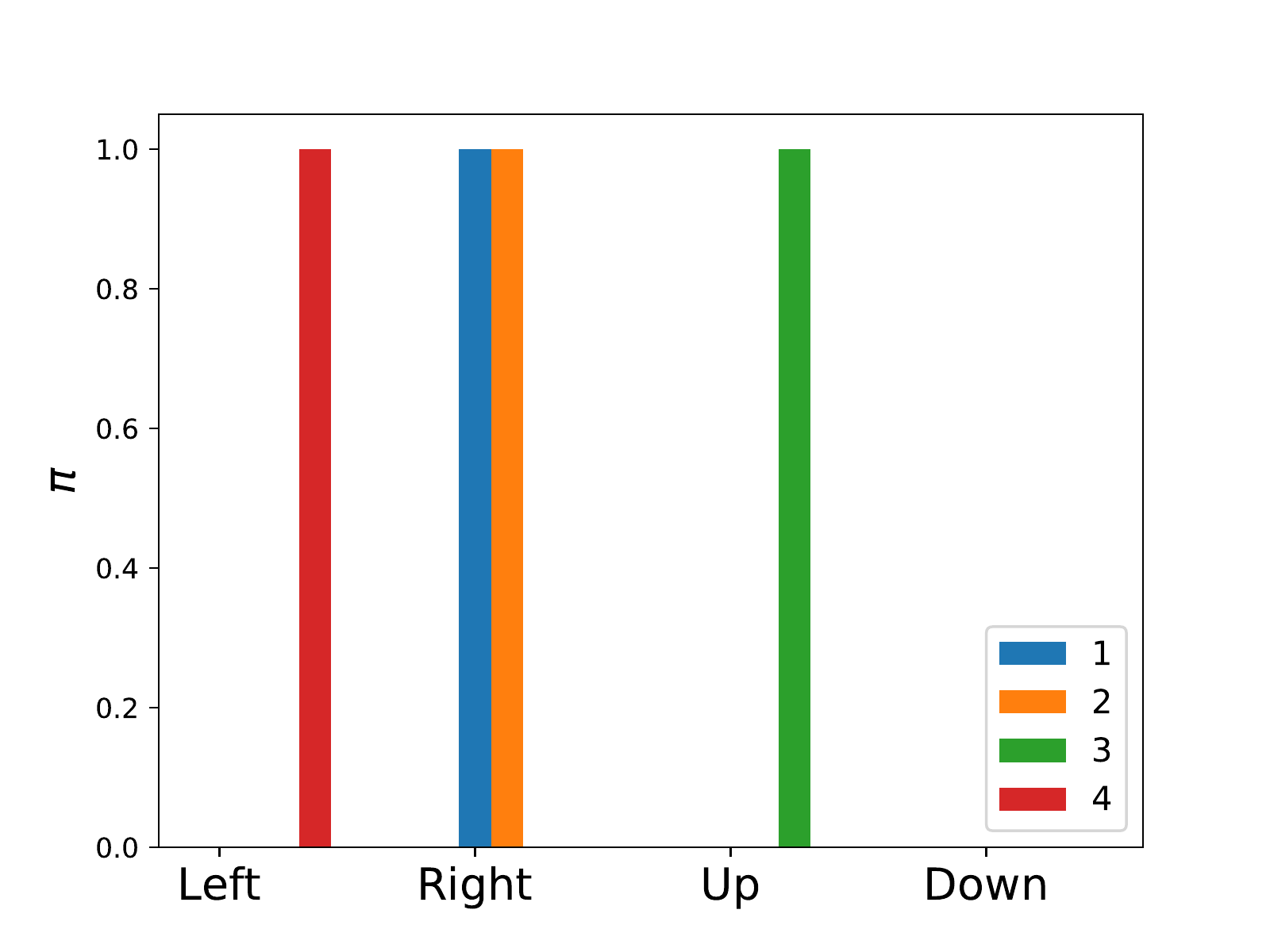}\\
(B)
\end{center}
\end{minipage}
\begin{minipage}{0.3\hsize}
\begin{center}
\includegraphics[width=1.0\linewidth]{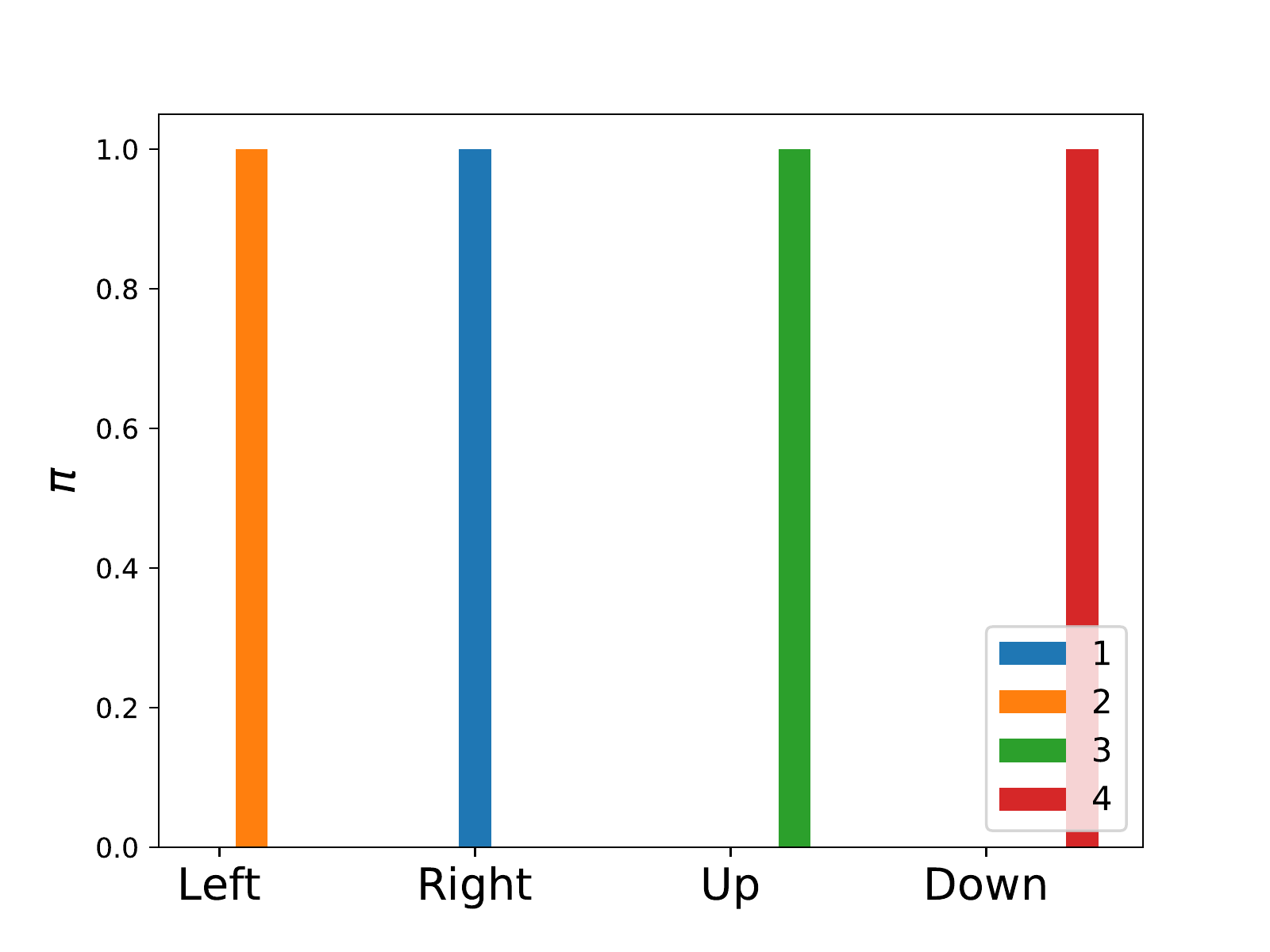}\\
(C)
\end{center}
\end{minipage}
\begin{minipage}{0.3\hsize}
\begin{center}
\includegraphics[width=1.0\linewidth]{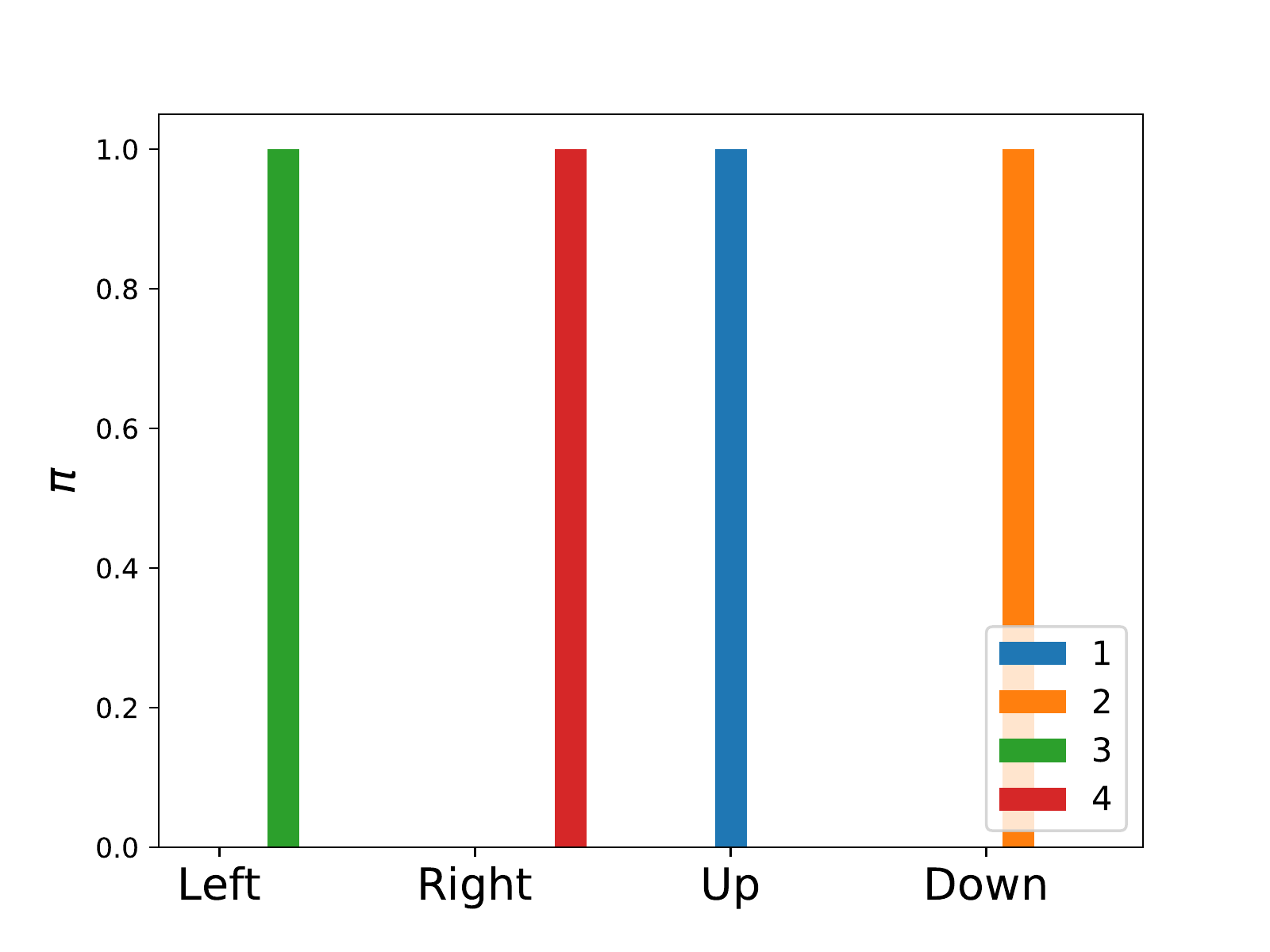}\\
(D)
\end{center}
\end{minipage}
\begin{minipage}{0.3\hsize}
\begin{center}
\includegraphics[width=1.0\linewidth]{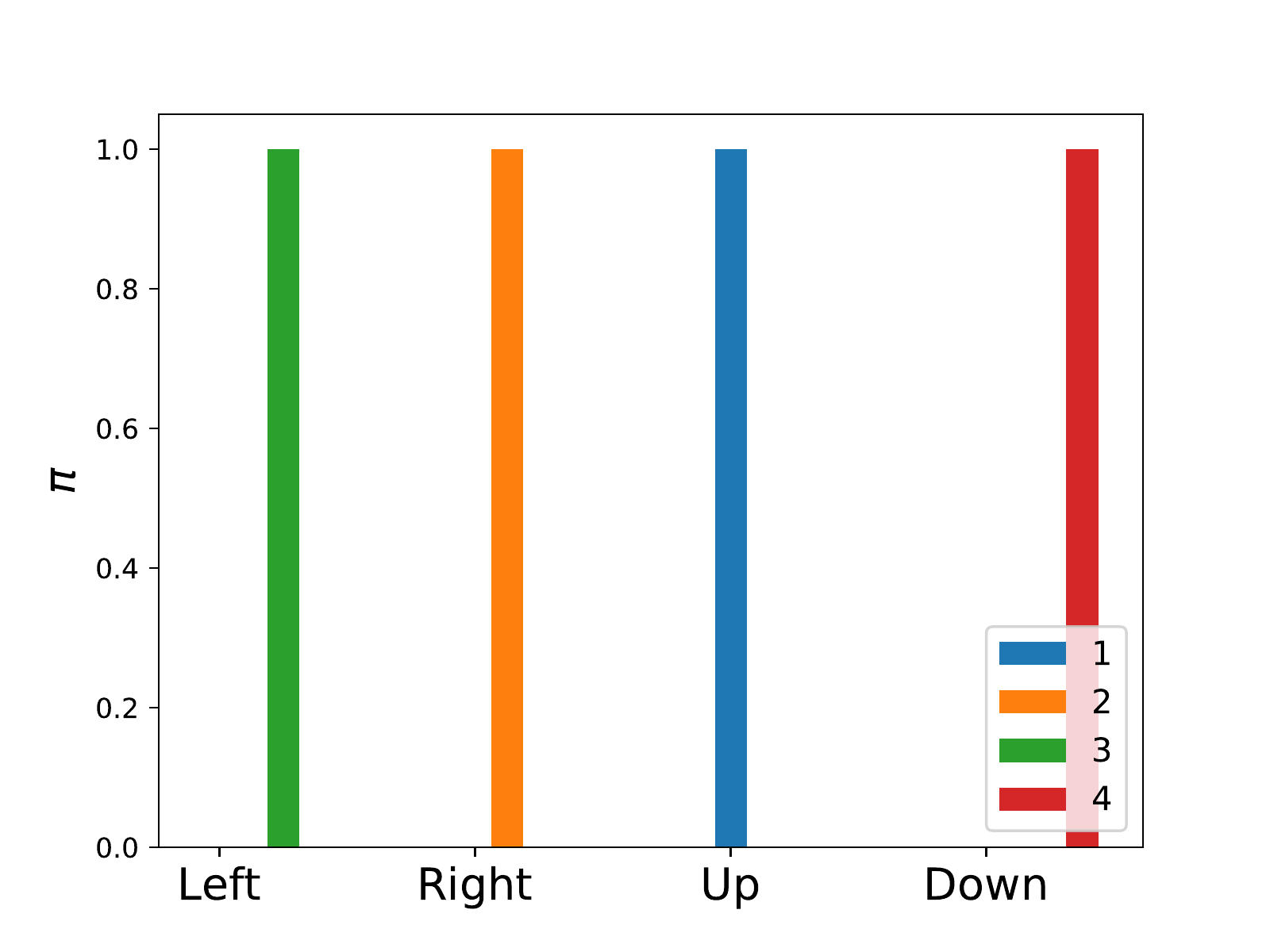}\\
(E)
\end{center}
\end{minipage}
\begin{minipage}{0.3\hsize}
\begin{center}
\includegraphics[width=1.0\linewidth]{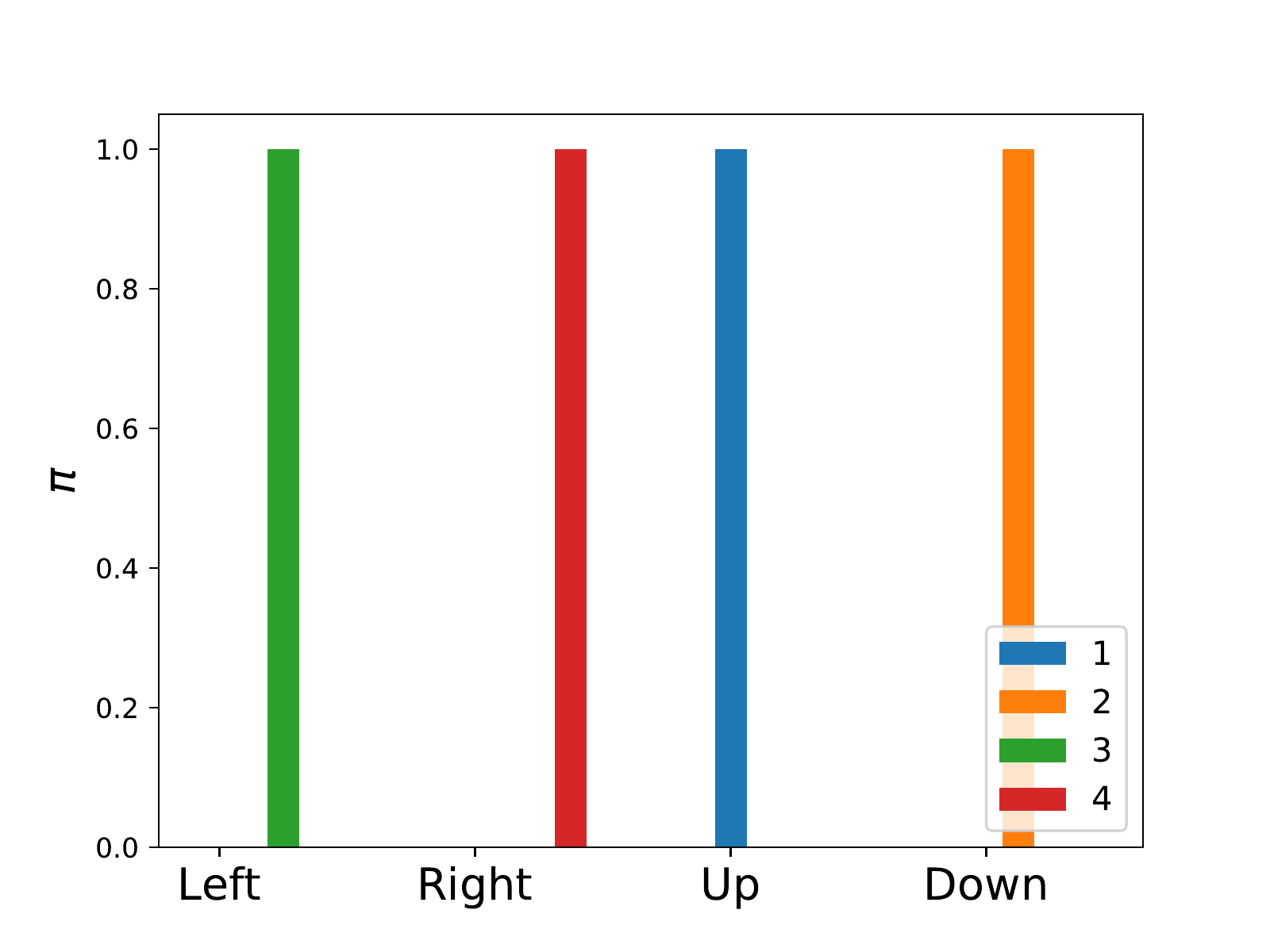}\\
(F)
\end{center}
\end{minipage}
\caption{Policies for the DNN of the controllable object. (A)-(C) show the $\pi_{\psi_k}$ of the Thomas' model, a model without pretraining, and model with pretraining for situation 1, respectively, and (D)-(F) show $\pi_{\psi_k}$ for situation 2. The order is the same as situation 1. Color represents the policy's number: blue is $\pi_{\psi_1}$, orange is $\pi_{\psi_2}$, green is $\pi_{\psi_3}$, and red is $\pi_{\psi_4}$.}
\label{fig:without}
\end{figure}

Here, we evaluate a correlation between latent features and the controllable object coordinates and policy $\pi_{\phi_k} (. \mid f(\vect{x}))$.

The correlation coefficients between the $k$-th neuron of $f_c$, i.e., $f_{c,k}$, and the controllable object's coordinates for situations 1 and 2 are given in tables~\ref{tbl:correlation} and \ref{tbl:correlation_4}.
As shown in table~\ref{tbl:correlation}, $f_{c,1}$ and $f_{c,2}$ are strongly correlated with the $x$-coordinate, whereas $f_{c,3}$ and $f_{c,4}$ are strongly correlated with the $y$-coordinate.
On the other hand, table~\ref{tbl:correlation_4} shows the opposite correlation, i.e., $f_{c,1}$ and $f_{c,2}$ are strongly correlated with the $y$-coordinate, and $f_{c,3}$ and $f_{c,4}$ are strongly correlated with the $x$-coordinate.
These strong correlations indicate a strong linearity; and we can locate the controllable object using $f_{c,k}(\vect{x})$.
In these tables, values in parentheses indicate the difference between the results obtained with the Thomas' model and the proposed model.
These values indicate that the Thomas' model exhibits linearity between $f_{k}(\vect{x})$ and the controllable object's coordinates even when the environment includes obstacles.
The results demonstrate that neuron $f_k(\vect{x})$ trained with the Thomas' model reacts to one of the controllable object's actions, whereas it is easy to reconstruct only the controllable object with $g(f(\vect{x}))$.

Table~\ref{tbl:correlation_without} shows the correlation without pretraining. 
The value to the left of the slash is situation 1 and the value to the right is situation 2.
As shown in situation 1, each neuron is not strongly correlated with the $y$-coordinate.
On the other hand, each neuron is strongly correlated with corresponding coordinates in situation 2, although the controllable object is reconstructed as part of the uncontrollable obstacles~(Fig.~\ref{fig:reconst_4_objects}~(B)).

Figure~\ref{fig:without} shows the policy values.
Here, trained policy is $\pi_{\phi_k, a} \approx 1.0$ when $f_{c,k}$ shows the strong correlation. 
In addition, the sign of the change of the controllable object's coordinate~(i.e., plus if the action is {\sf right} and {\sf down}; otherwise, minus) is the same as the sign of the correlation. 
However, differing from other results, Fig.~\ref{fig:without}~(B) shows that $\pi_{\phi_k, {\sf right}}\approx 1.0, (k=1,2)$ and $\pi_{\phi_k, {\sf down}} \approx 0.0, (\forall k)$.
These results also indicate that pretraining is effective for stable training.

\subsection{Analysis of DNN for Uncontrollable Obstacles}

\begin{figure}[t]
\centering
\begin{minipage}{0.35\hsize}
\begin{center}
\includegraphics[width=1.0\linewidth]{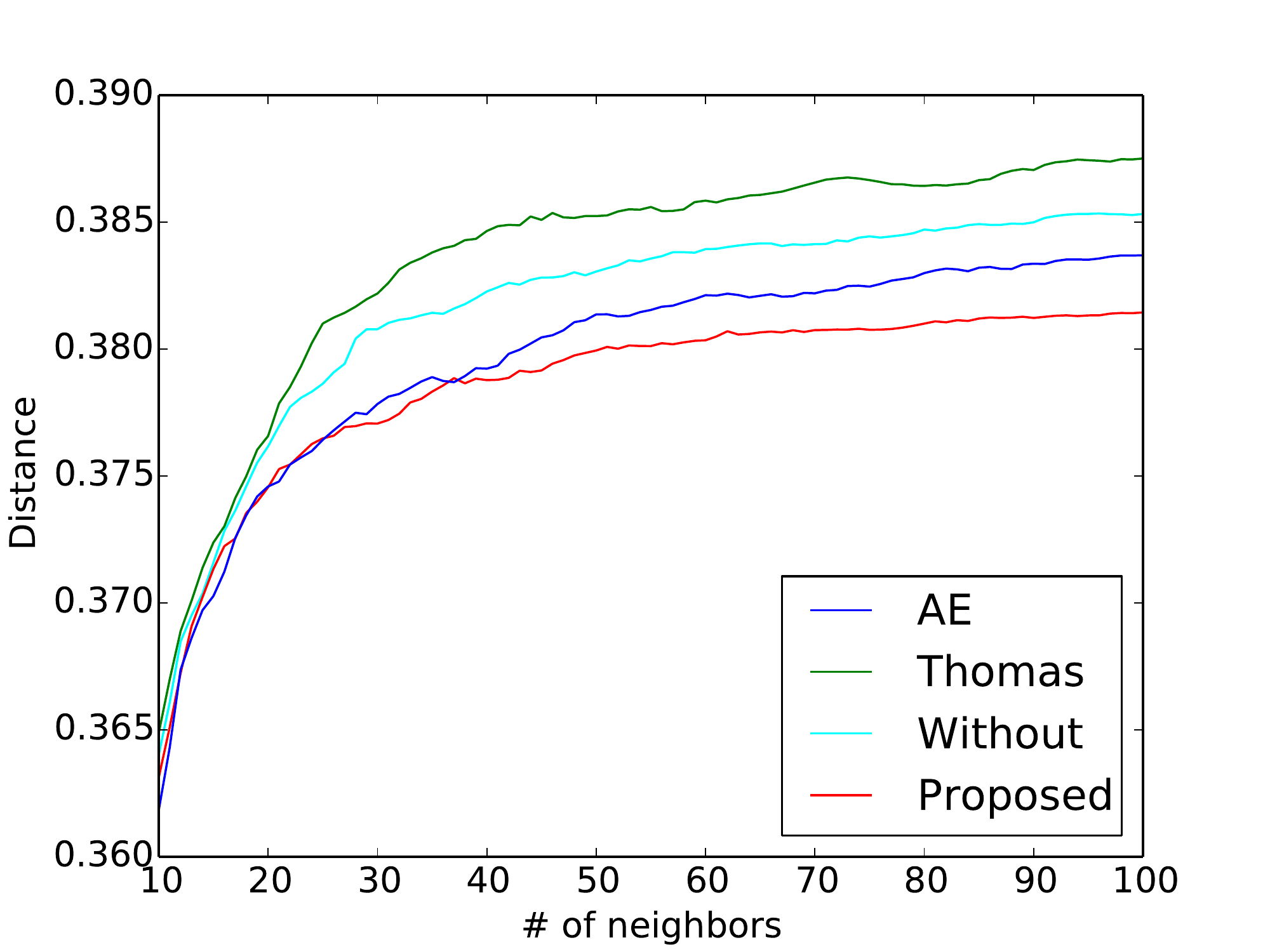}\\
Situation 1
\end{center}
\end{minipage}
\begin{minipage}{0.35\hsize}
\begin{center}
\includegraphics[width=1.0\linewidth]{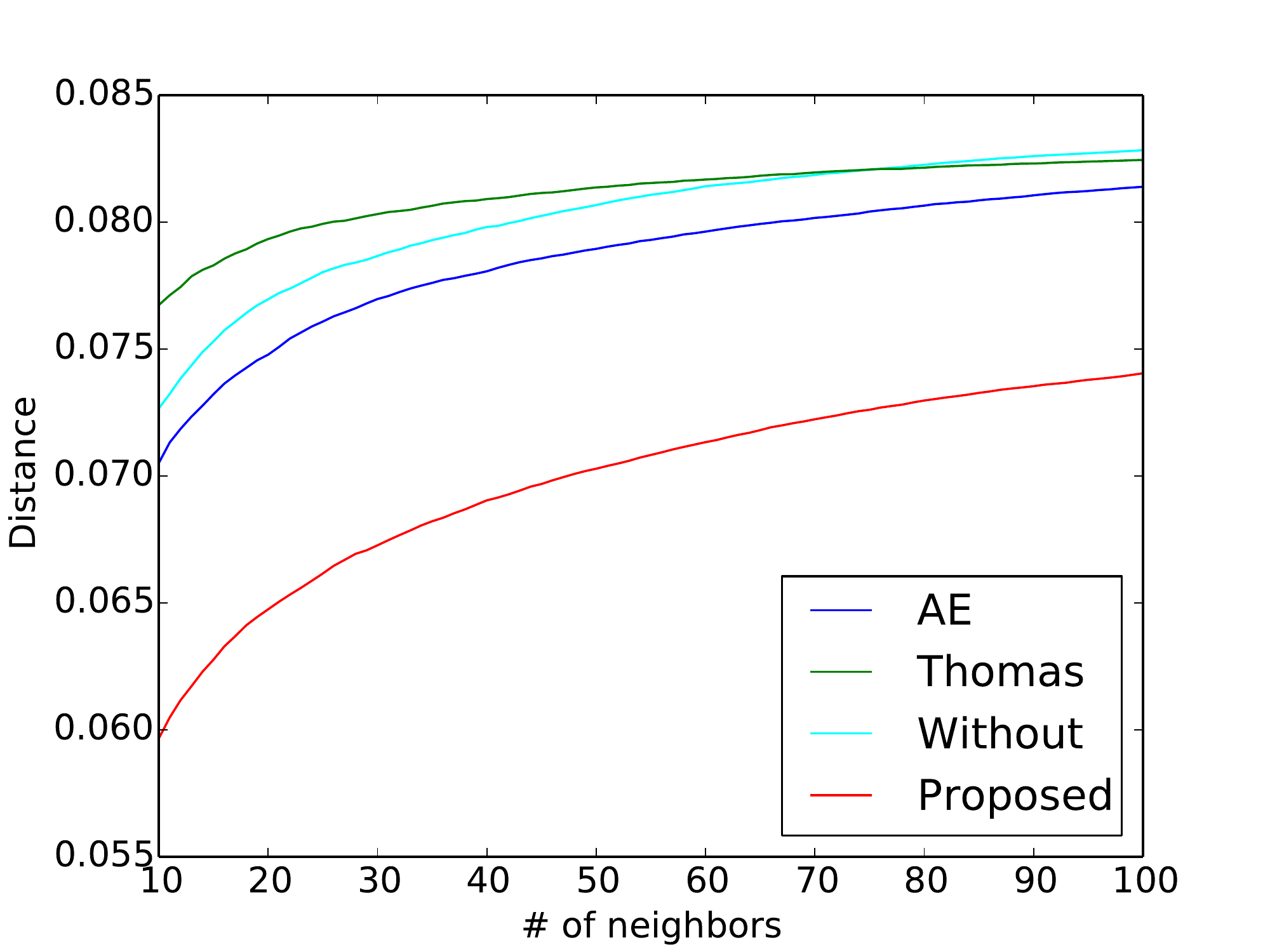}\\
Situation 2
\end{center}
\end{minipage}
\caption{Averaged accumulated Euclidean distance between obstacle's coordinates in $\vect{x}_i$ and $j$-th neighborhood $\vect{x}_i^j$. $\vect{x}_i^j$ is computed in the latent space. Red represents $f_u$ of the proposed model~(Proposed), cyan represents $f_u$ of the model without pretraining~(Without), and blue and green represent the results of the autoencoder~(AE) and the Thomas' model~(Thomas). }
\label{fig:distance}
\end{figure}

\begin{figure}[t]
\centering
\begin{minipage}{0.35\hsize}
\begin{center}
\includegraphics[width=1.0\linewidth]{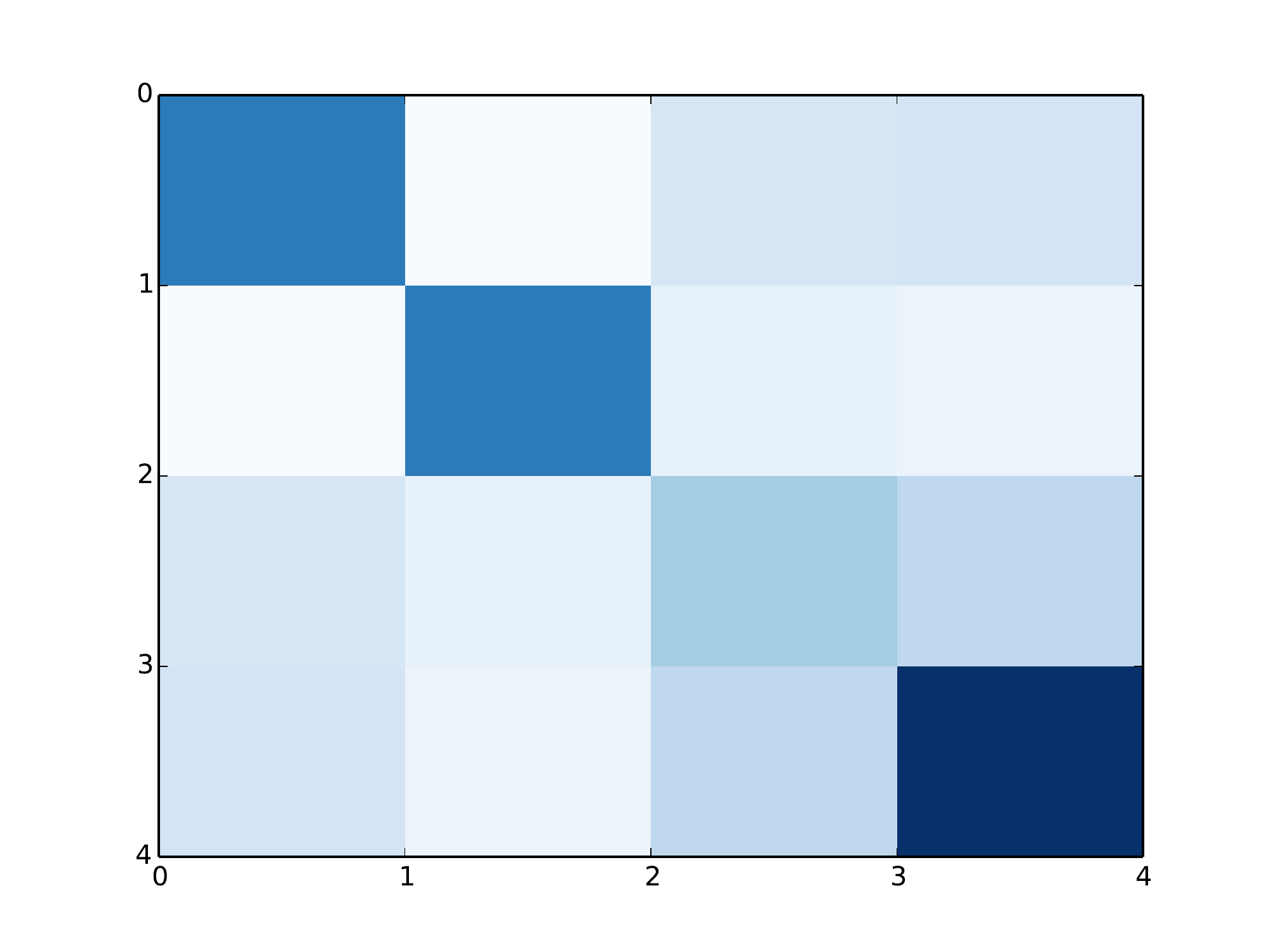}\\
Situation 1
\end{center}
\end{minipage}
\begin{minipage}{0.35\hsize}
\begin{center}
\includegraphics[width=1.0\linewidth]{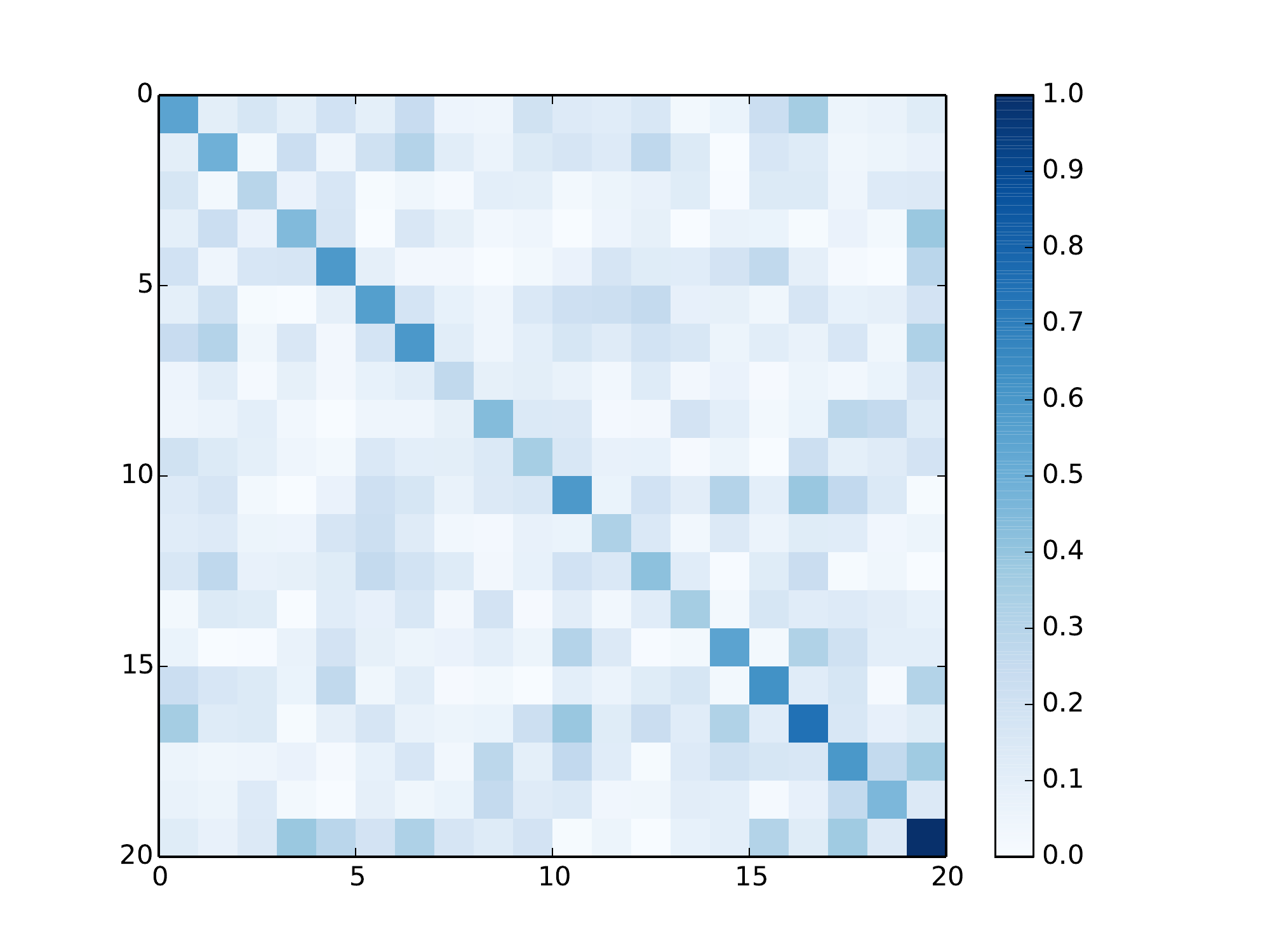}\\
Situation 2
\end{center}
\end{minipage}
\caption{Normalized absolute concentration matrices computed by $f_u(\vect{x})$.}
\label{fig:inverse}
\end{figure}

In Section~\ref{sub:controllable}, we presented the analysis of $f_c(\vect{x})$ for the controllable object. 
In this section, we analyze other, i.e., $f_u(\vect{x})$ for uncontrollable obstacles.

Figure~\ref{fig:distance} shows the averaged accumulated Euclidean distances between obstacle's coordinates in $\vect{x}_i$ and the $j$-th neighborhood $\vect{x}_i^j$, where neighbor $\vect{x}_i^j$ is computed in the latent space. 
As can be observed, the distance of the latent features $f_u$ of the proposed pretraining model is less than that of other models to the closeness when the obstacles coordinate in the input images.

Figure~\ref{fig:inverse} shows normalized absolute concentration matrices computed using $f_u$ of the model with pretraining.
It is known that the value of a concentration matrix is dependent between variables.
As can be seen, the variables of $f_u(\vect{x})$ are independent.

These results indicate that the proposed model can construct the independent and meaningful latent feature $f_u(\vect{x})$; however, it is unclear whether this ability works well in any environment.
In a future study, we plan to investigate the effectiveness of the proposed model in a more complex environment.

\subsection{RL Task involving Acquisition of Extrinsic Rewards}

We investigated the behavior of the proposed model with an RL task under a modified version of situation 1 that involved extrinsic rewards.
We applied the recurrent-based Q-network followed by our encoders.
Details about the environment and network architecture are described in the Supplementary Materials.
Note that the controllable object returns to the starting point if it hits the obstacle.

\begin{figure}[t]
\centering

\begin{minipage}{0.3\hsize}
\begin{center}
\includegraphics[width=1.0\linewidth]{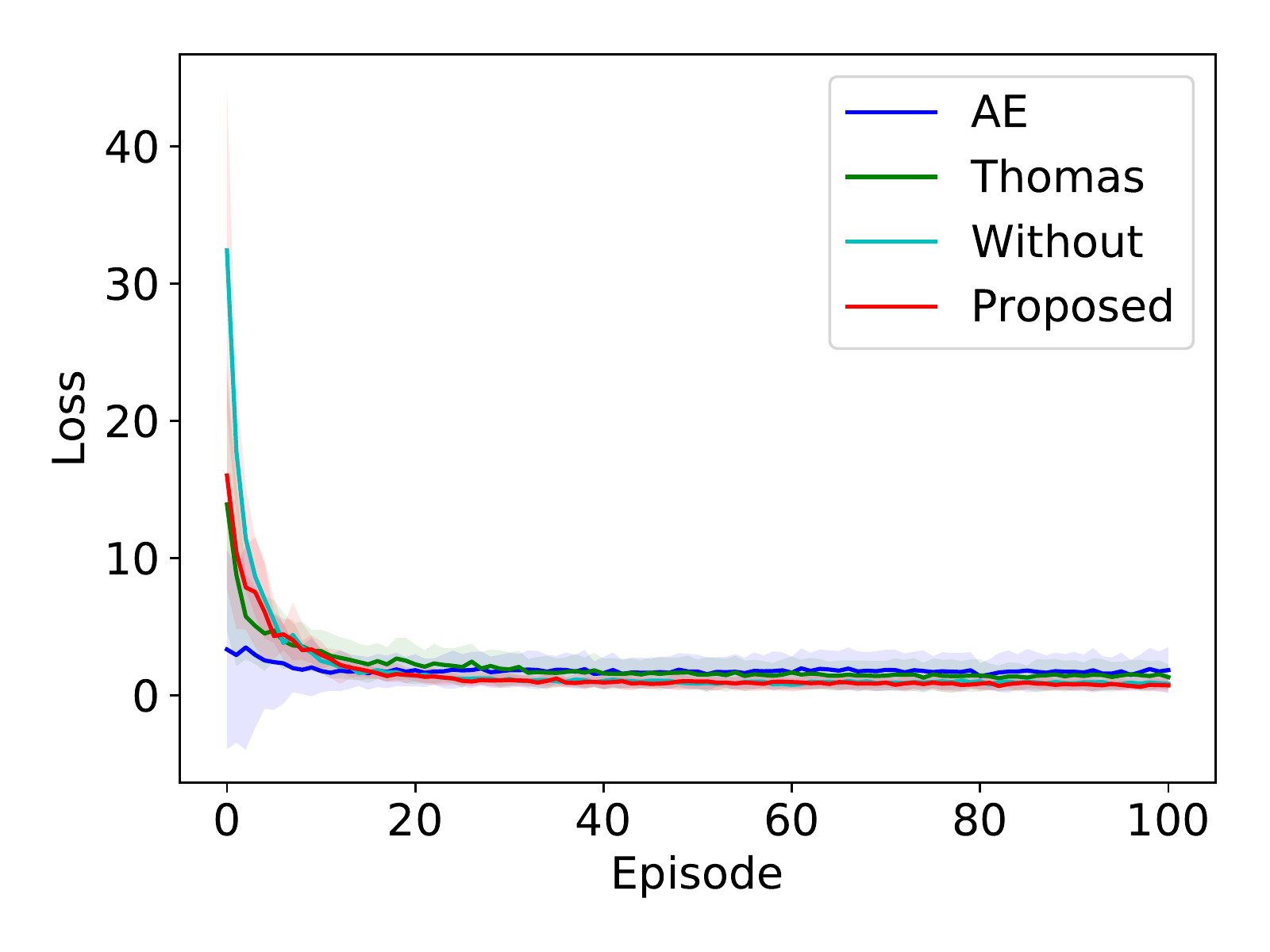}\\
(A)
\end{center}
\end{minipage}
\begin{minipage}{0.3\hsize}
\begin{center}
\includegraphics[width=1.0\linewidth]{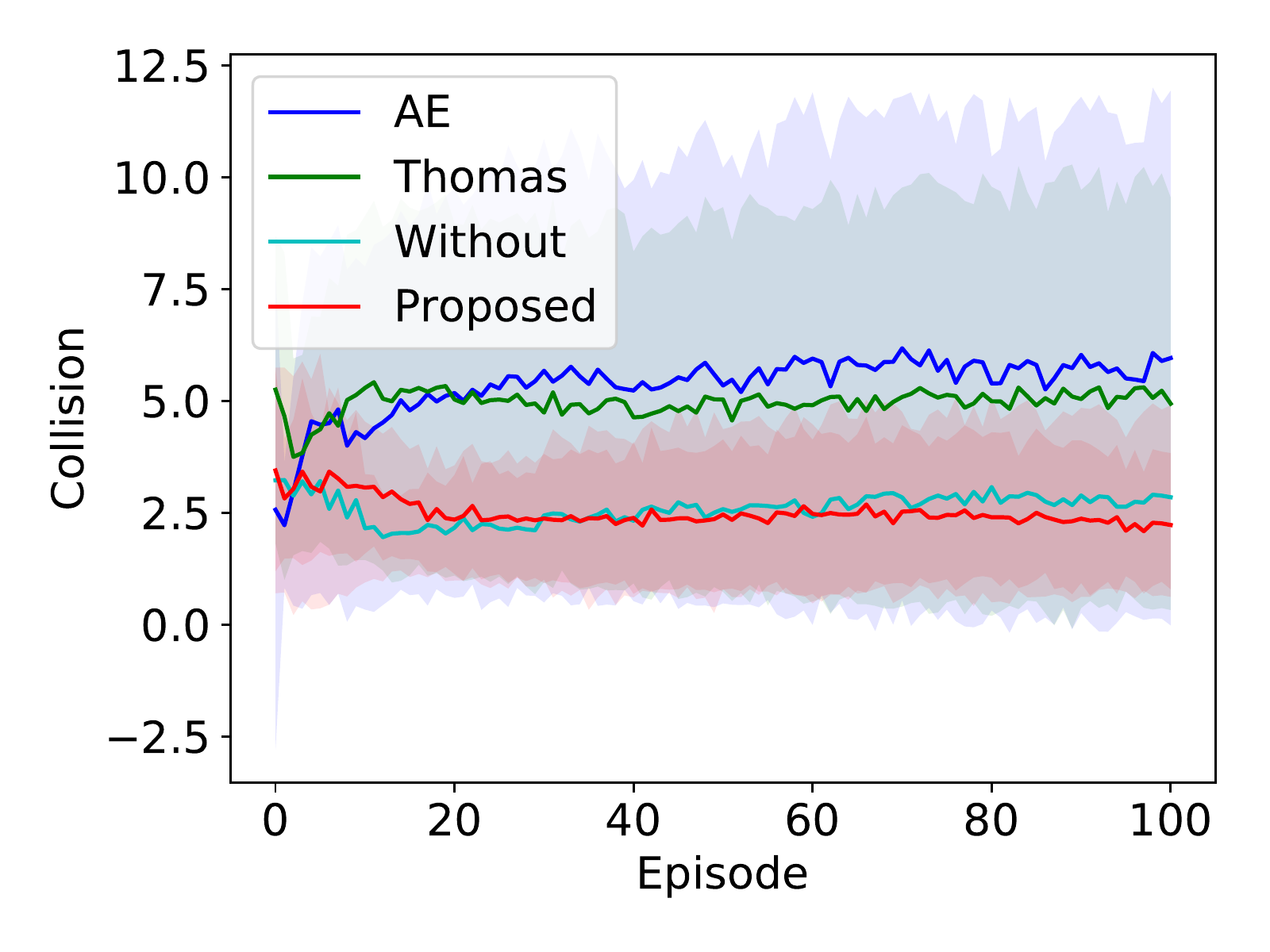}\\
(B)
\end{center}
\end{minipage}
\begin{minipage}{0.3\hsize}
\begin{center}
\includegraphics[width=1.0\linewidth]{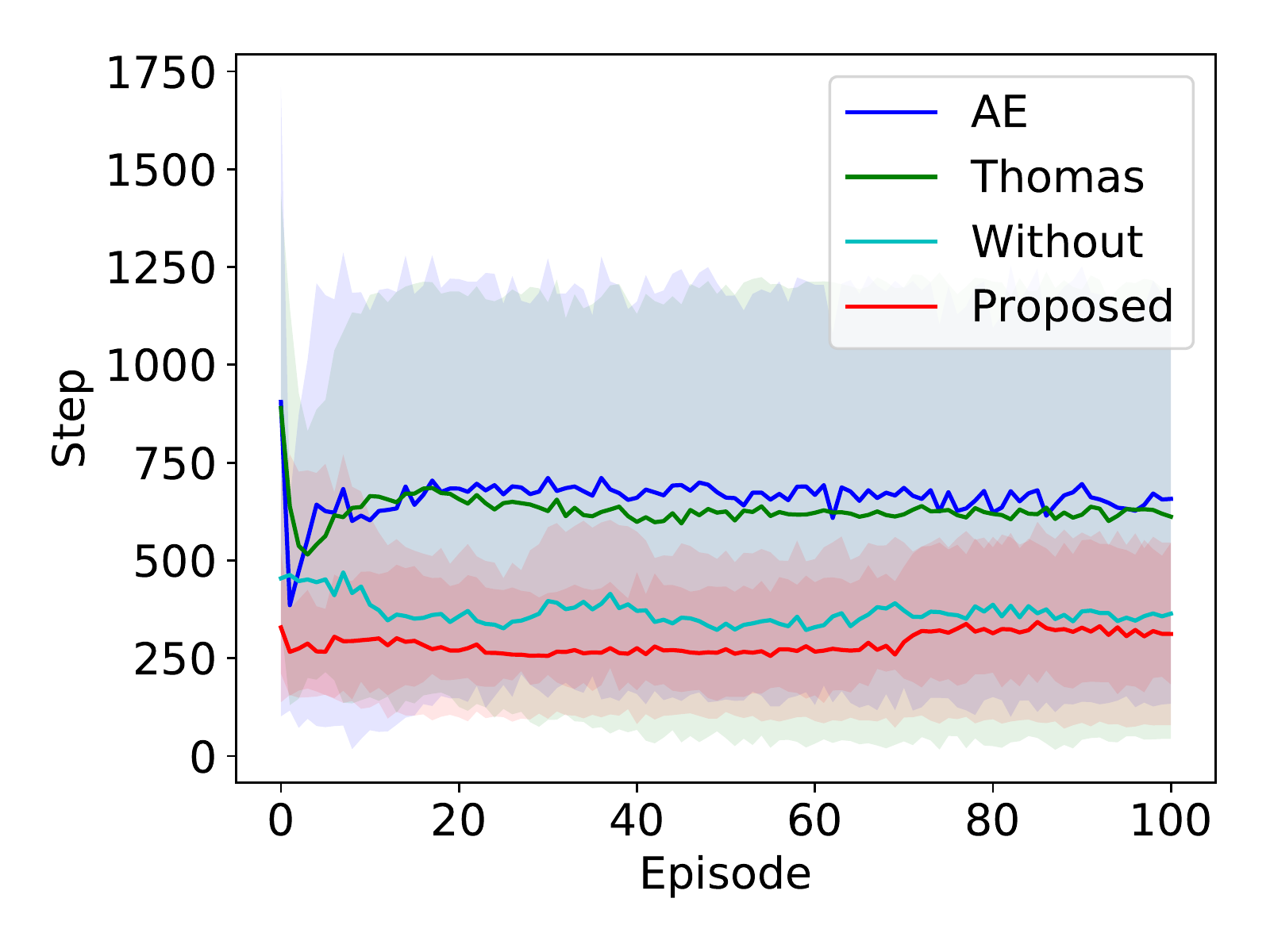}\\
(C)
\end{center}
\end{minipage}
\caption{Results of a task with extrinsic rewards. (A)-(C) represent the losses of the Q-network, numbers of collisions, and numbers of steps required to reach the goal, respectively. In each case, we conducted 10 trials and computed the average (lines) and variance (shaded areas). The results obtained using the autoencoder~(AE), the Thomas' model~(Thomas), the model without pretraining~(Without) and the proposed model~(Proposed) are shown in blue, green, cyan, and red, respectively.}
\label{fig:simu_rl}
\end{figure}

Figure~\ref{fig:simu_rl} shows the results for the autoencoder, the Thomas' model, the model without pretraining, and the proposed model.
Figure~\ref{fig:simu_rl}~(A) shows that all models converged, and Figs.~\ref{fig:simu_rl}~(B) and (C) show that the proposed method required fewer steps than the other two methods.
The autoencoder can represent the controllable and uncontrollable objects, whereas latent features were entangled.
In addition, the Thomas' model removes the information of the obstacle, and the model without pretraining cannot disentangle correctly.
These results indicate the possibility that the proposed model is effective for an RL task with sparse extrinsic rewards.

\section{Conclusion}
In this article, we introduce a method to disentangle independently controllable and uncontrollable factors of variation without annotated data.
We analyzed the model proposed by Thomas et al.~\cite{thomas2017independently,thomas2017independently2} when the target environment included the uncontrollable obstacles.
Although the Thomas' method can disentangle independent controllable factors of variation, it has no mechanism to reconstruct obstacles, which is an issue the proposed method addresses by building upon their approach.
The proposed method solves this problem by training two DNNs simultaneously, i.e., one that represents the controllable object and another that represents uncontrollable obstacles.
For training, we used the parameters obtained by training the Thomas' model as our initial parameters to focus on the controllable object.

We evaluated the proposed method by conducting simulations and found that it can disentangle independently controllable and uncontrollable factors of variation.
We have also shown that the previous model ignores uncontrollable obstacles and two DNNs without pretraining could not disentangle correctly.

A limitation of the current study is a lack of experimental results for a wider range of environments.
Furthermore, it may be possible to train $f_u$ and $g_u$ to be more independent and meaningful. 
In a future study, we plain to investigate the extensibility of the proposed method and how easily it can be extended to more complex environments.



\section*{Supplementary Materials}

\setcounter{section}{1}
\def\thesection{\Alph{section}}

\subsection{Derivation of Objective Function for Controllable Ojbect}
As describe in~\cite{thomas2017independently2}, ${\cal S}_{i,k}$ has many possible variations.
In this article, we derived ${\cal S}_{i,k}$ from a following equation which links the lower bound of the mutual information ${\cal L}(\vect{\varphi}, f \mid f') \geq \sup \mathbb{E}_{p(\varphi \mid f)} \left[ {\cal S}_{\varphi} \right]$~\cite{thomas2017independently}.
\begin{eqnarray}
{\cal S}_{\varphi} = \mathbb{E} \left[ \log \left( \frac{A(f (\vect{x}), f(\vect{x}'), \vect{\varphi})}{\mathbb{E}_{p(\vartheta \mid f)} [A(f(\vect{x}), f(\vect{x}'), \vect{\vartheta})]} \right) \middle| s' \sim P_{s,s'}^{\pi_{\phi_{\varphi}}}\right]
\end{eqnarray}
In this article, we set $A(f,f',\vect{\vartheta})=\mathbb{E}_{p(\vartheta \mid f)} [\hat{A}] + \hat{A}$. Then,
\begin{eqnarray}
{\cal S}_{ \varphi} &=& 
\mathbb{E}_{s'} \left[ \log \left( \frac{\mathbb{E}_{p(\vartheta \mid f)} [\hat{A}(f(\vect{x}), f(\vect{x}'), \vect{\vartheta})] + \hat{A}(f(\vect{x}), f(\vect{x}'), \vect{\varphi})}{\mathbb{E}_{p(\vartheta \mid f)} \left[\mathbb{E}_{p(\vartheta \mid f)} [\hat{A}(f(\vect{x}), f(\vect{x}'), \vect{\vartheta})] + \hat{A}(f(\vect{x}), f(\vect{x}'), \vect{\vartheta}) \right]} \right)
\right] \nonumber \\
&=& \mathbb{E}_{s'} \left[ \log \left( \frac{\mathbb{E}_{p(\vartheta \mid f)} [\hat{A}(f(\vect{x}), f(\vect{x}'), \vect{\vartheta})] + \hat{A}(f(\vect{x}), f(\vect{x}'), \vect{\varphi})}{2\mathbb{E}_{p(\vartheta \mid f)} \left[ \hat{A}(f(\vect{x}), f(\vect{x}'), \vect{\vartheta}) \right]} \right)
\right] \nonumber \\
&=& \mathbb{E}_{s'} \left[ \log \frac{1}{2} \left( 1 + \frac{\hat{A}(f (\vect{x}), f'(\vect{x}'), \vect{\varphi})}{\mathbb{E}_{p(\vartheta \mid f)} [\hat{A}(f(\vect{x}), f(\vect{x}'), \vect{\vartheta})]} \right)
\right], 
\end{eqnarray}
where $\vect{\varphi}$ (and $\vect{\vartheta}$) represents controllable factors of variation~\cite{thomas2017independently}, $\hat{A}$ represents a score describing how close vector $\vect{\varphi}$ is to the variation it caused in ($f(\vect{x}), f(\vect{x}'_j)$).
We used $\mathbb{E}[\mathbb{E}[X]] = \mathbb{E}[X]$ to expand from the top equation to the 2nd equation.
In this article, we approximate the expectation $\mathbb{E}$ as the summation as follows.
\begin{eqnarray}
{\cal S}_{\varphi} \approx \tilde{{\cal S}}_{\varphi} = \frac{1}{KN} \sum_a \pi_{{\phi}_{\varphi}} (a \mid f(\vect{x})) \sum_{s'} \log \frac{1}{2} \left( 1 + \frac{\hat{A}(f(\vect{x}), f(\vect{x}'), \vect{\varphi})}{\mathbb{E}_{p(\vartheta \mid f)} [\hat{A}(f(\vect{x}), f(\vect{x}'), \vect{\vartheta})]} \right),
\label{equ:sub_sup}
\end{eqnarray}
where $N$ represents the number of samples sampled under $P^{\pi_{{\psi}_{\varphi}}}_{s, s'}$.
We set $\hat{A} = \| f(\vect{x}')-f(\vect{x}) \|^\top \vect{\varphi}$ where $\vect{\varphi}$ as a one-hot vector ($k$-th variable $\varphi_k$ is $1$ and otherwise $0$)~\cite{thomas2017independently2}, $p(\varphi \mid f) = 1/K$, and $\{ \vect{\vartheta}_k \mid k = 1, 2, \cdots, K \}$ where $\vect{\vartheta}_k$ is the one-hot vector which $k$-th variable is $1$~($\vect{\varphi} = \vect{\vartheta}_k$). 
We substitute them into above equation and get
\begin{eqnarray}
\tilde{{\cal S}}_{\varphi} &=& \frac{1}{KN} 
\sum_a \pi_{{\phi}_{\varphi}} (a \mid f(\vect{x}))
\sum_{s'} \log \frac{1}{2} \left( 1 + \frac{ K \mid f(\vect{x}') - f(\vect{x})\mid^\top \vect{\vartheta}_k}{\sum_{k'} \mid f(\vect{x}') - f(\vect{x})\mid^\top \vect{\vartheta}_{k'}} \right) \\
&=& \frac{1}{KN} \sum_{a} \pi_{\psi_k}(a \mid f(\vect{x})) \sum_{s'} \log \left( \frac{1}{K} +  \frac{\mid f_{k}(\vect{x}')-f_{k}(\vect{x}) \mid}{\sum_{k'} \mid f_{k'}(\vect{x}')-f_{k'}(\vect{x}) \mid} \right) + \gamma,
\label{equ:sub}
\end{eqnarray}
where $\gamma$ is a constant because $\sum_a \pi_{\psi_k}(a \mid f(\vect{x}))=1$.
After removing $\gamma$ and adding the reconstruction error, we can get our objective function for the controllable object. 
\begin{eqnarray}
\argmin_{\vect{Z}} \mathbb{E}_{s} \left[ {\cal R} - \frac{\lambda'}{KN} \mathbb{E}_{p(\varphi  \mid f)} [\tilde{{\cal S}}_{\varphi}] \right]  \approx \argmin_{\vect{Z}} \sum_i {\cal R}_i - \lambda \sum_k {\cal S}_{i,k},
\end{eqnarray}
where $\vect{Z} = \{ \vect{\phi}, \vect{\theta}, \vect{\Psi}\}$, $\lambda'$ is hyperparameter, and $\lambda = \lambda'/KN$.
By using this equation, we can avoid $\log(0)$ computation due to $1/K$.

\subsection{Environment}
\label{sub:env}

\begin{figure}[t]
\centering
\includegraphics[width=0.25\linewidth]{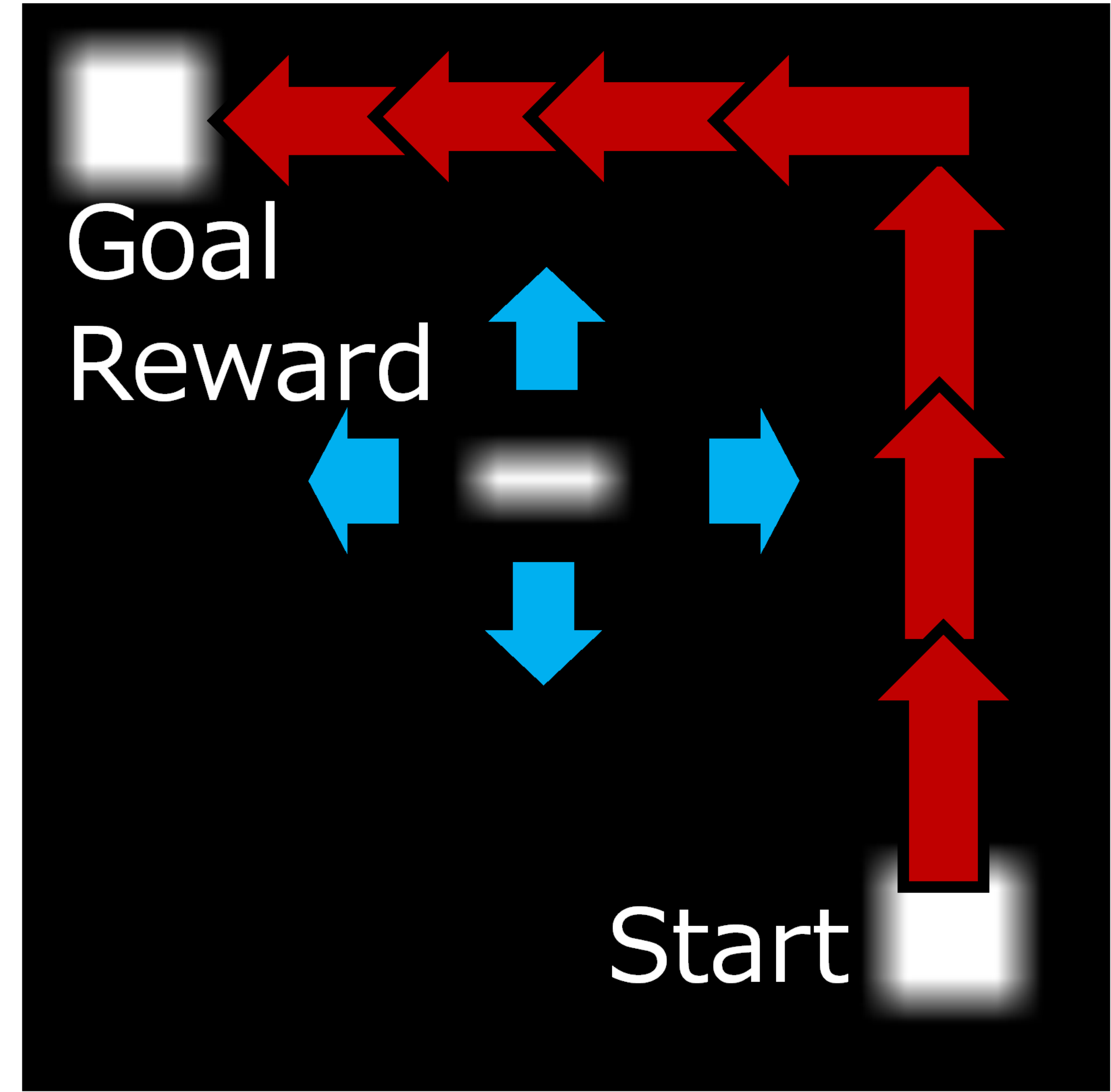}\\
\caption{Environment for extrinsic reward task. The squares represents controllable object, and the line object represents an obstacle.}
\label{fig:env_rewards}
\end{figure}

\begin{figure}[t]
\centering
\includegraphics[width=0.65\linewidth]{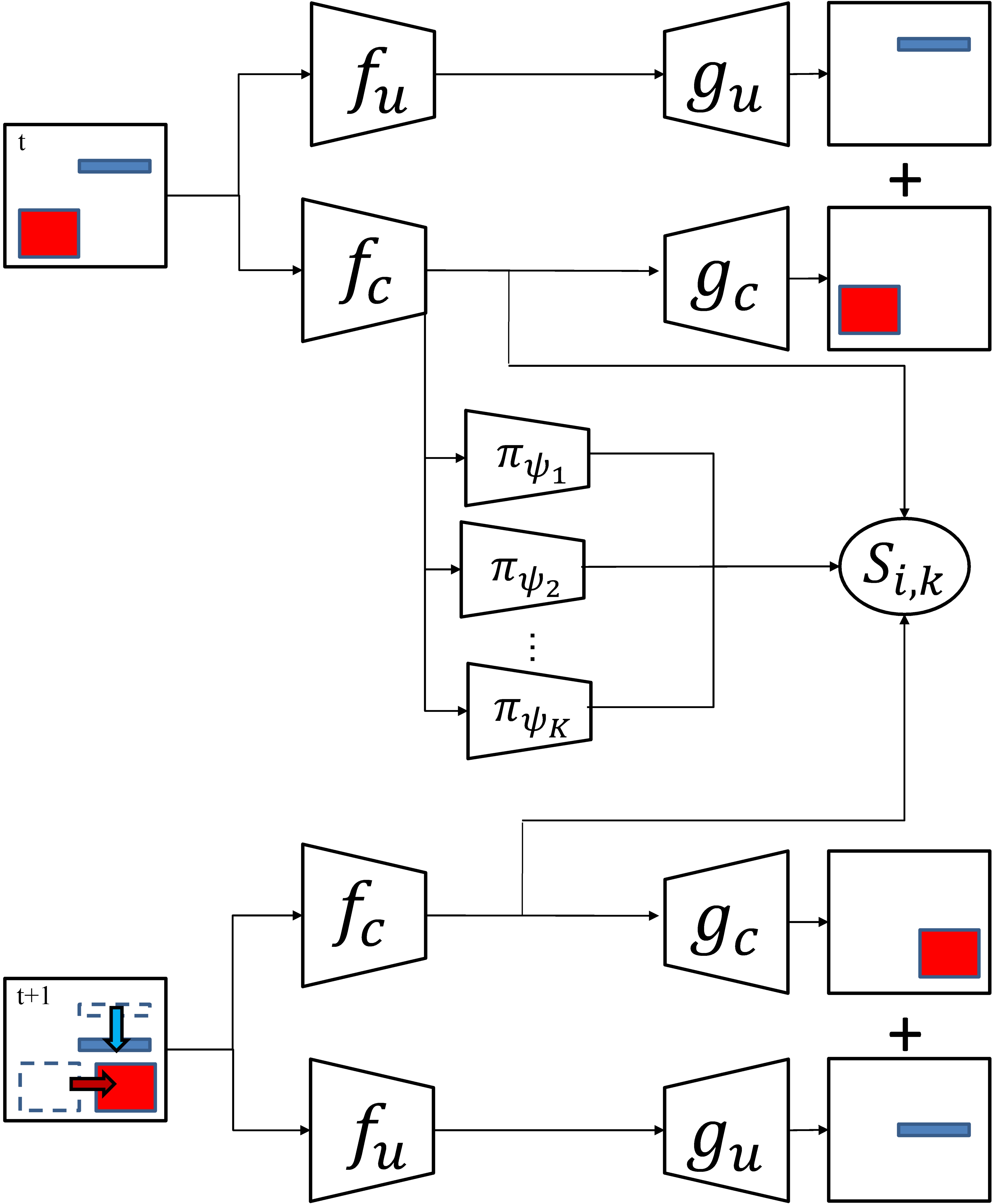}\\
\caption{Network architectures for disentangling the controllable and uncontrollable factors of variation.}
\label{fig:model1}
\end{figure}

\begin{figure}[t]
\centering
\includegraphics[width=0.65\linewidth]{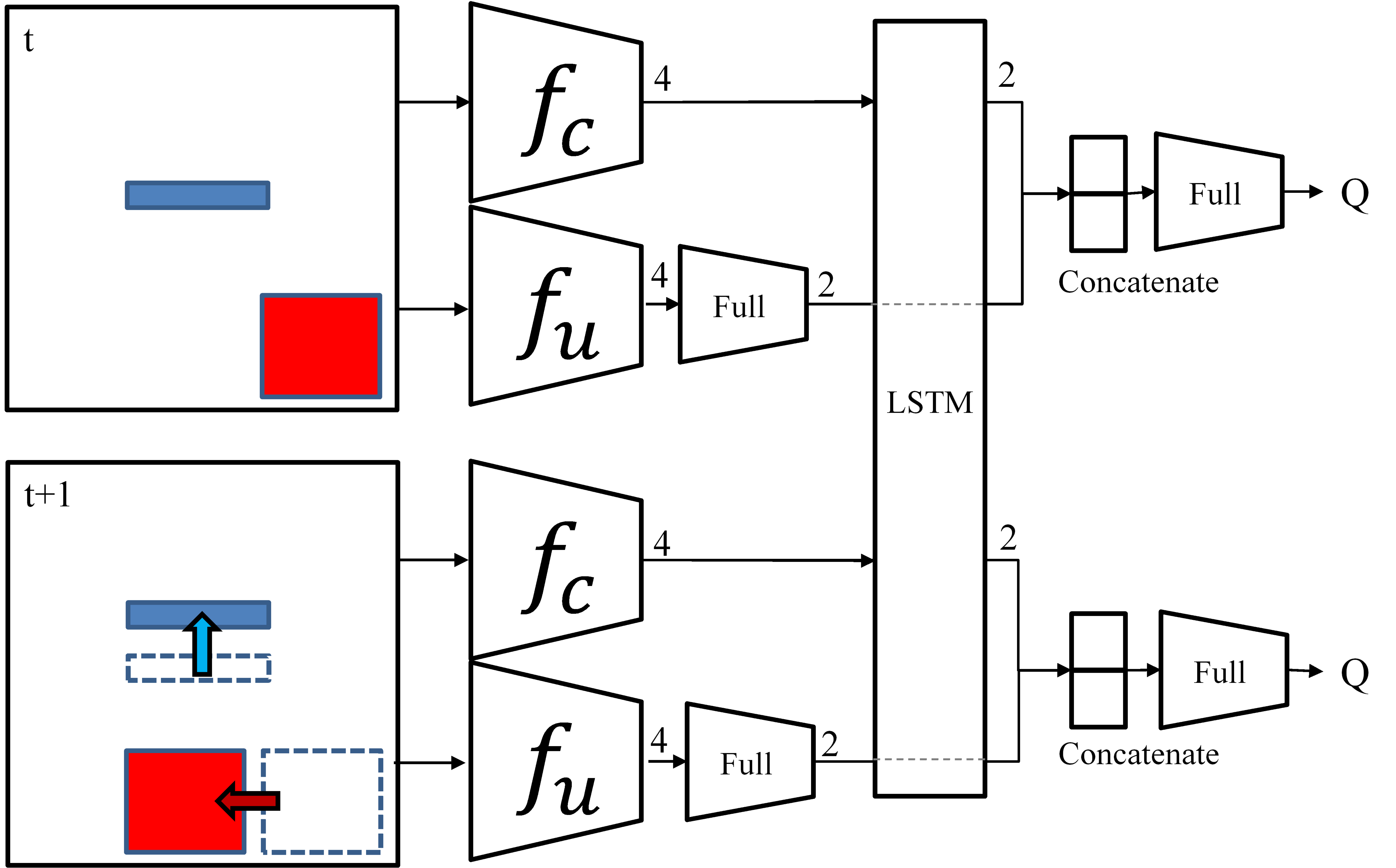}\\
\caption{Network architectures for the modified task with extrinsic rewards. Figures represent the number of dimenstion. Since the uncontrollable obstacles moves randomly, the proposed model feeds the uncontrollable latent features into fully connected layers, rather than the LSTM layer.}
\label{fig:model2}
\end{figure}

In situation 1, the red arrow represents a $3 \times 3$ controllable object, and the blue arrow represents a $3 \times 1$ uncontrollable object in $24 \times 24$-pixel images.
In situation 2, the red arrow represents a $3 \times 3$ controllable object, and the blue arrows represent a $3 \times 1$, $1 \times 3$, and five-dot uncontrollable objects in $24 \times 24$-pixel images.
All obstacles in the situation 2 move only within each quadrant.
If the controllable object hits the uncontrollable obstacles, it returns to the starting point $(20,20)$ in the situation 1 and $(2,2)$ in the situation 2.

All objects move randomly by up to three pixels along its action direction. 
We set $N=20$ and obtained a total of $200000$ training data from the environments.
Relative to correlation, averaged accumulated Euclidean distance~(Section~\ref{sub:distance}), and concentration matrix, computations were performed using $1000$ randomly sampled states from the environment.

We also applied the proposed model to a modified version of the task with sparse extrinsic rewards shown in Fig.~\ref{fig:env_rewards}, i.e., a reward of $+1$ for covering $(2,2)$ (and $0$ otherwise).
As same as the situation 1, if the controllable object hits the uncontrollable object, it returns to the starting point $(20,20)$.
If the controllable object reaches the goal, or if the number of steps reaches $2000$, each episode is terminated.

\subsubsection{Averaged Accumulated Euclidean Distance}
\label{sub:distance}
This section explains how to compute the averaged accumulated Euclidean distance.
Let $\vect{O}_l(\vect{x})$ denote $l$-th obstacle's coordinate vector in $\vect{x}$, $S$, $L$, $J$ denote the number of samples sampled from environment, the number of obstacles, and the number of neighbors, respectively.
Then, the averaged accumulated Euclidean distance $D$ is 
\begin{equation}
D = \frac{1}{SLJ} \sum_{i,l,j} \| \vect{O}_l(\vect{x}_i) - \vect{O}_l(\vect{x}_i^j) \|,
\end{equation}
where $\vect{x}_i^j$ represents $j$-th neighborhood computed in the latent space.

\subsection{Network Architectures}
Here, we explain the network architectures for disentangling the controllable and uncontrollable factors of variation and the task that involves the acquisition of extrinsic rewards.
In all cases, the hyperparameters of the model without pretraining were set to the same values as the proposed model, and the Thomas' model and the autoencoder were the same as the controllable and uncontrollable object models, respectively.

\subsubsection{Disentangling Network}
The architecture of our disentangling network is as follows.
The controllable object's encoder $f_{c}$ comprised $16 \times 4 \times 4$ and $16 \times 3 \times 3$ ReLU convolutional layers with a stride of 2, followed by a fully connected ReLU layer with $32$ units and a tanh layer with $K = 4$ features.
The architecture of the controllable object's decoder $g_{c}$ was the inverse of this architecture.
Here, the hyperparameters were taken from the literature~\cite{bengio2017independently}.
The architectures for the uncontrollable object, i.e., $f_{u}$ and $g_{u}$, were identical to $f_{c}$ and $g_{c}$ in the environment shown in situation 1.
On the other hand, in situation 2, the number of dimensions of the highest hidden layer of $f_{u}$ was set to $20$.
Here, the policy (weight) $\pi_{\psi_k}$ was a softmax function over $K$ actions computed from the output of the fully connected layer. 
Figure~\ref{fig:model1} shows an overview of the proposed model.

We used the Adam optimizer and REINFORCE estimator and used setting $\lambda = 0.05$ in all cases.
We adjusted $\lambda$ in each situation such that the Thomas' model maximized the number of reconstructed objects while disentangling controllable factors of variation.
Meanwhile, we used the default values provided by chainer~\cite{tokui2015chainer} for the other hyperparameters.

\subsubsection{Network for Handling Extrinsic Rewards}

\begin{figure}[t]
\centering
\includegraphics[width=0.25\linewidth]{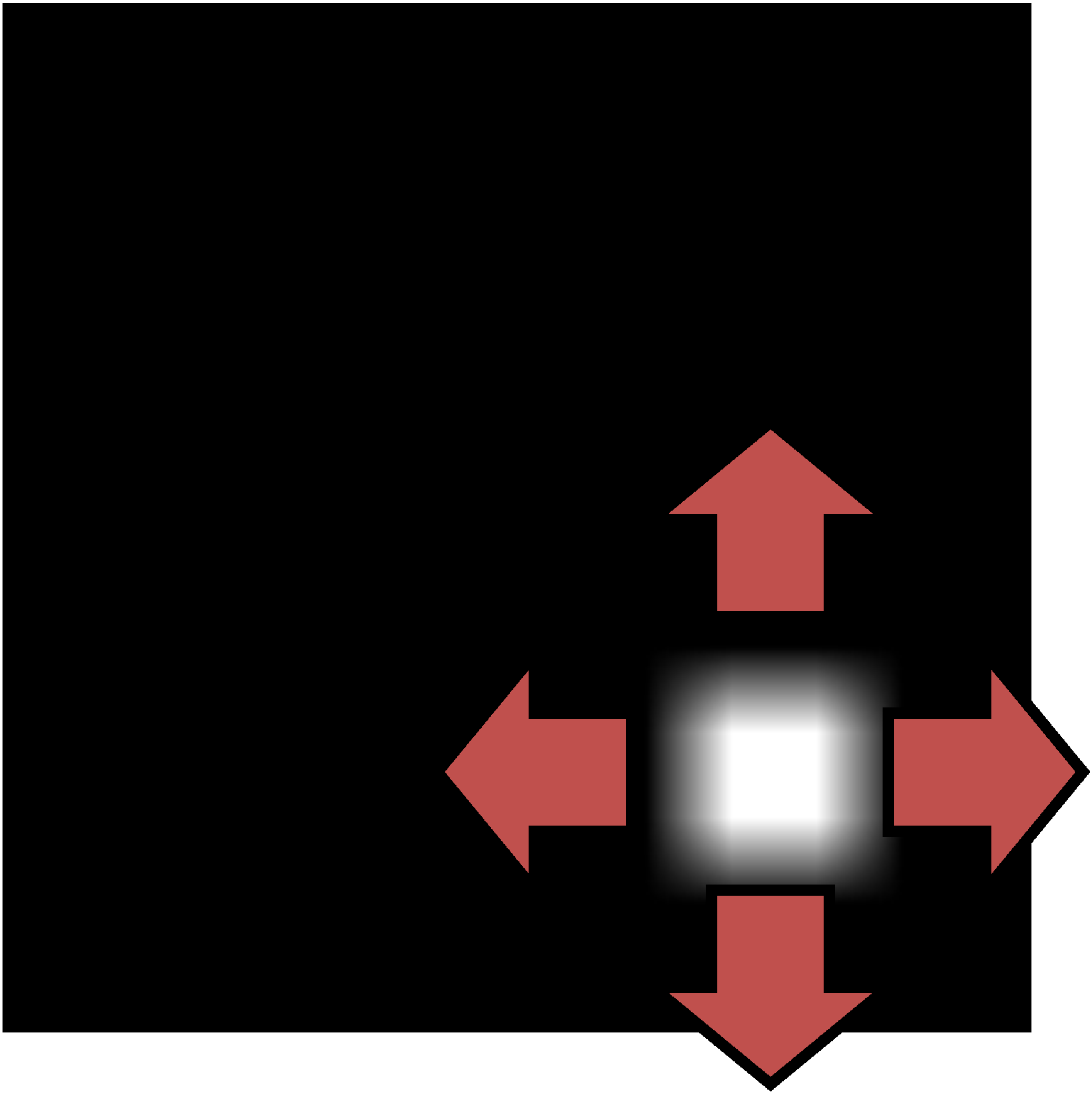}\\
\caption{Simple environment with only a controllable object.}
\label{fig:env}
\end{figure}

\begin{figure}[t]
\centering
\begin{minipage}{0.33\hsize}
\begin{center}
\includegraphics[width=1.0\linewidth]{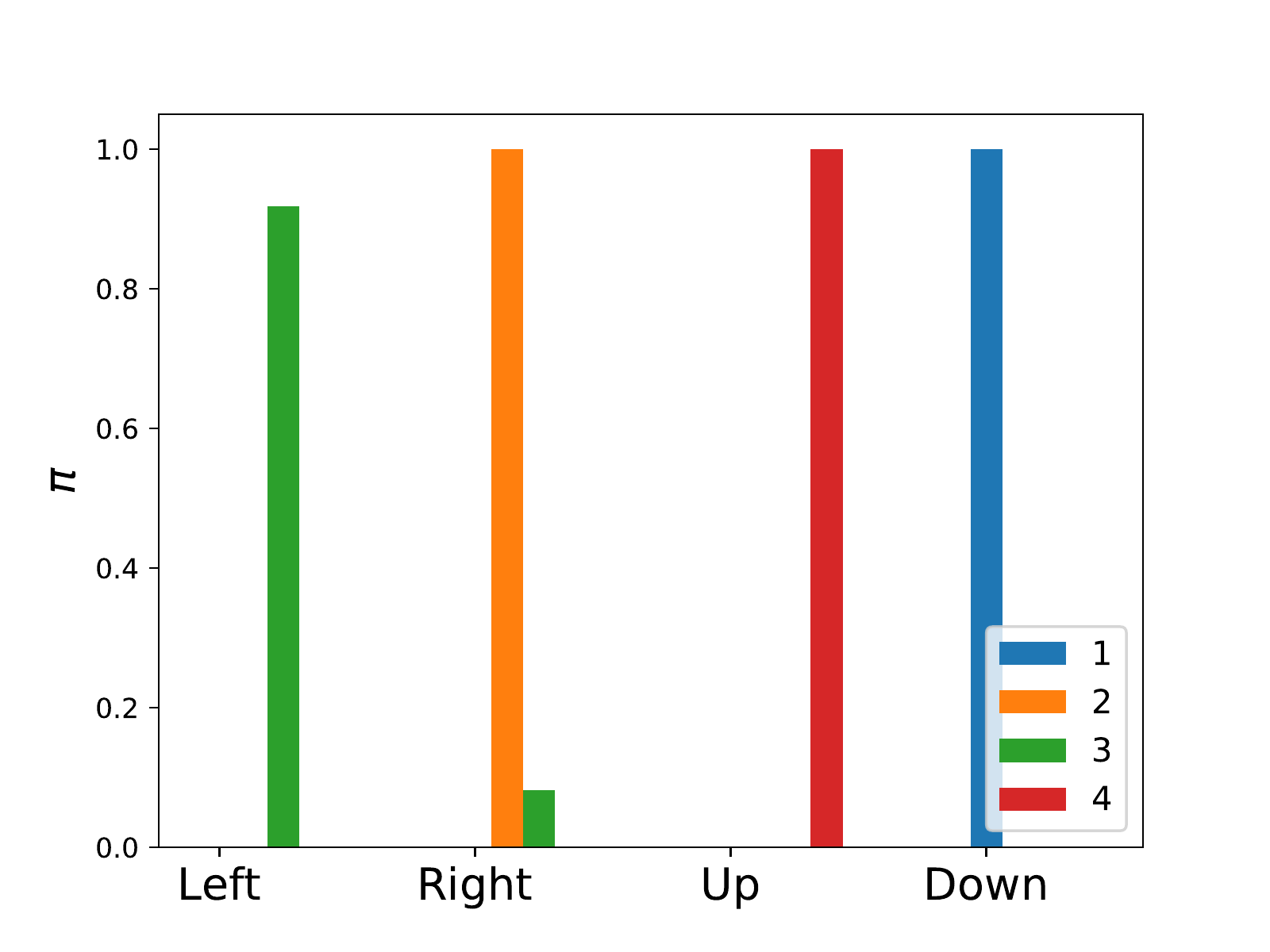}\\
Situation 1
\end{center}
\end{minipage}
\begin{minipage}{0.33\hsize}
\begin{center}
\includegraphics[width=1.0\linewidth]{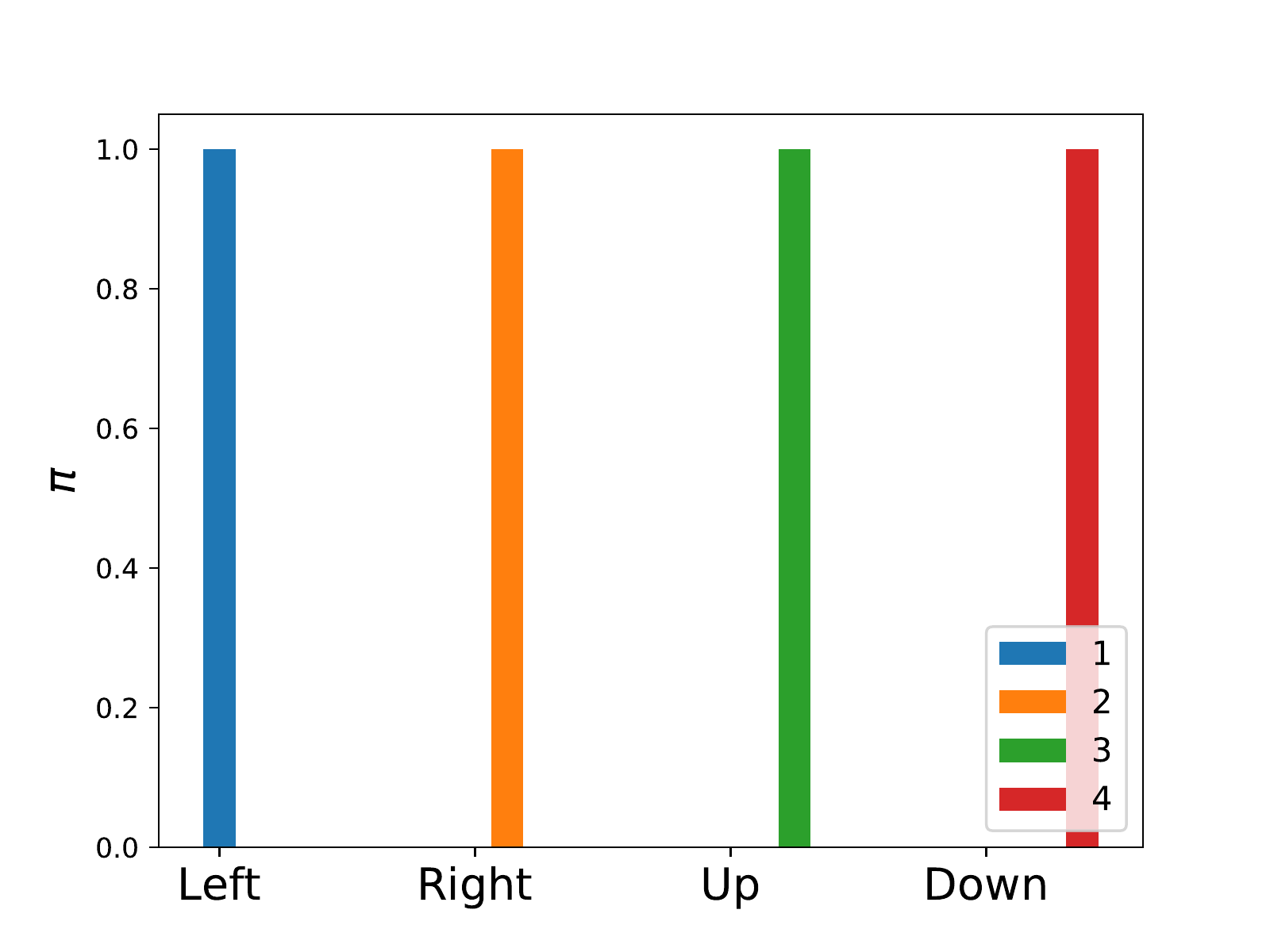}\\
Situation 2
\end{center}
\end{minipage}
\caption{Policies of pretrained model in simple environment.}
\label{fig:weight}
\end{figure}

\begin{figure}[t]
\centering
\includegraphics[width=0.65\linewidth]{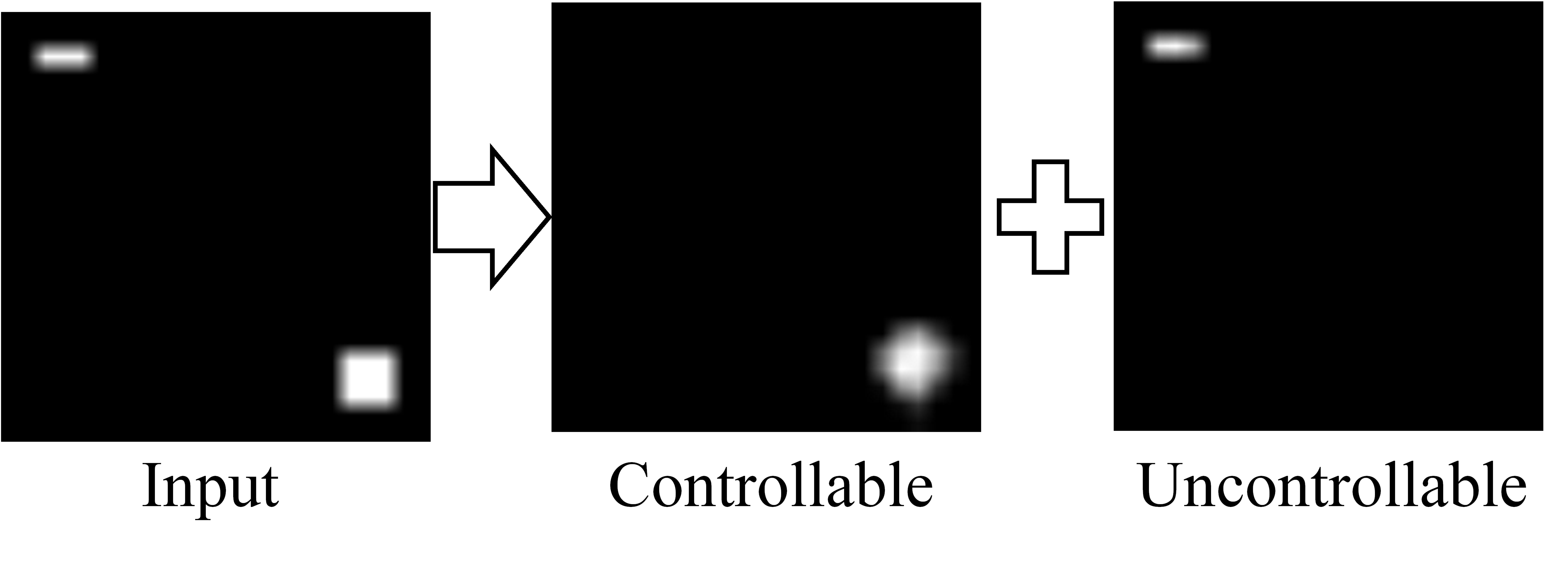}
\caption{Example reconstruction results of situation 1 when we reused parameters from the simple environment.}
\label{fig:reconst_reuse}
\end{figure}

\begin{figure}[t]
\centering
\includegraphics[width=0.65\linewidth]{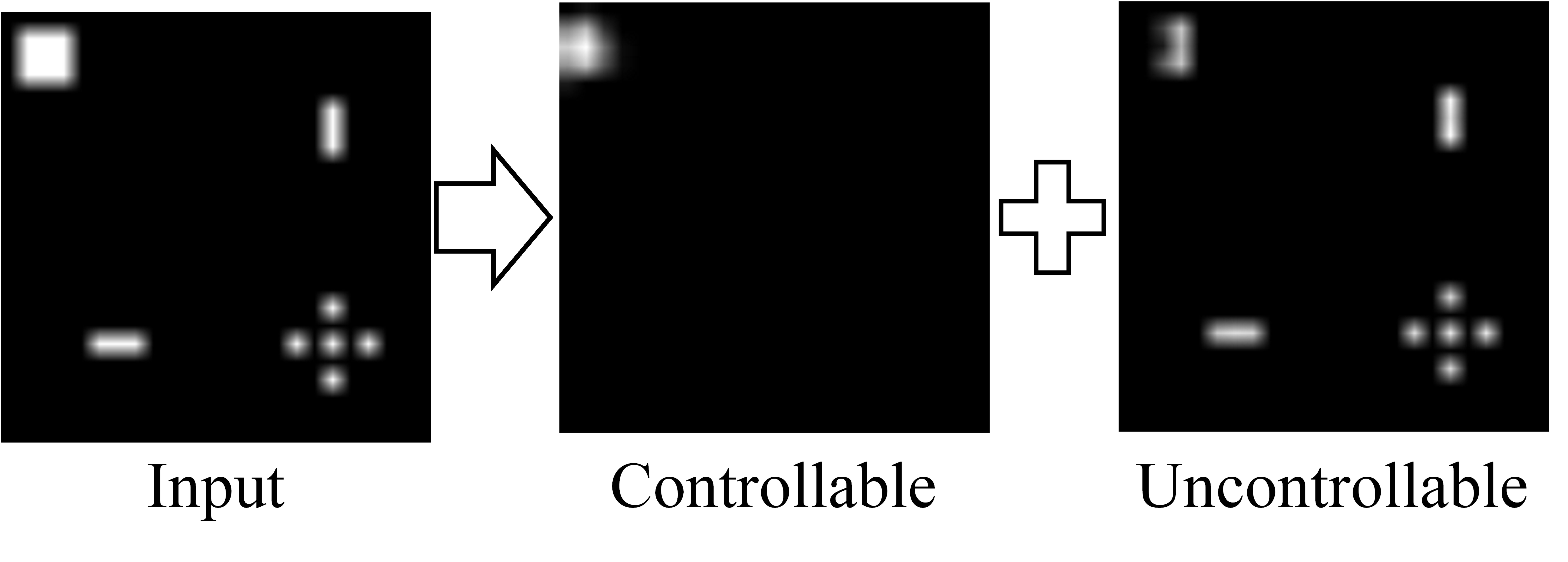}
\caption{Example reconstruction results of situation 2.}
\label{fig:reconst_reuse_miss}
\end{figure}

\begin{table}[t]
\begin{center}
\caption{Correlation coefficients of situation 1 when reusing parameters from the simple environment.}
\begin{tabular}{|l|c|c|c|c|}
\hline
Axis & $f_{c, 1}$ & $f_{c, 2}$ & $f_{c, 3}$ & $f_{c, 4}$ \\
\hline \hline
$x$ & $0.108$ & $0.904$ & $-0.816$ & $-0.018$\\
\hline
$y$ & $0.934$ & $-0.135$ & $0.167$ & $-0.759$\\
\hline
\end{tabular}
\label{tbl:correlation_reusing}
\end{center}
\end{table}

\begin{table}[t]
\begin{center}
\caption{Correlation coefficients of situation 2.}
\begin{tabular}{|l|c|c|c|c|}
\hline
Axis & $f_{c, 1}$ & $f_{c, 2}$ & $f_{c, 3}$ & $f_{c, 4}$ \\
\hline \hline
$x$ & $-0.846$ & $0.735$ & $-0.009$ & $-0.064$\\
\hline
$y$ & $-0.080$ & $-0.130$ & $-0.897$ & $0.742$\\
\hline
\end{tabular}
\label{tbl:correlation_reusing_miss}
\end{center}
\end{table}

We applied the proposed model to an RL task that included extrinsic rewards using a modified version of the deep recurrent Q-network~(DRQN)~\cite{hausknecht2015deep}.
As shown in Fig.~\ref{fig:model2}, the proposed model fed the controllable and uncontrollable latent features into long short-term memory~(LSTM) and fully connected layers, respectively.
It then concatenates these outputs and feeds the results into a further fully connected layer to compute the Q-values.
The LSTM layer and first fully connected layer each comprise two units, whereas the Q-value computation layer outputs $K$ values that correspond to the actions. 
We froze the two encoders and used stochastic gradient descent with a learning rate of $1.0 \times 10^{-3}$ and a discount factor of $0.9$.
Both the Thomas' model and the autoencoder were based on the original DRQN~(i.e., single route) and all hyperparameters were set to the same values in both cases.
Other networks can be available such as~\cite{mao2018universal,pathak2017curiosity}, however, this article does not mention the optimal network architecture.

\subsection{Additional Results}

\subsubsection{Example of Other Approaches for Stable Training}

\begin{figure}[t]
\centering
\begin{minipage}{0.35\hsize}
\begin{center}
\includegraphics[width=1.0\linewidth]{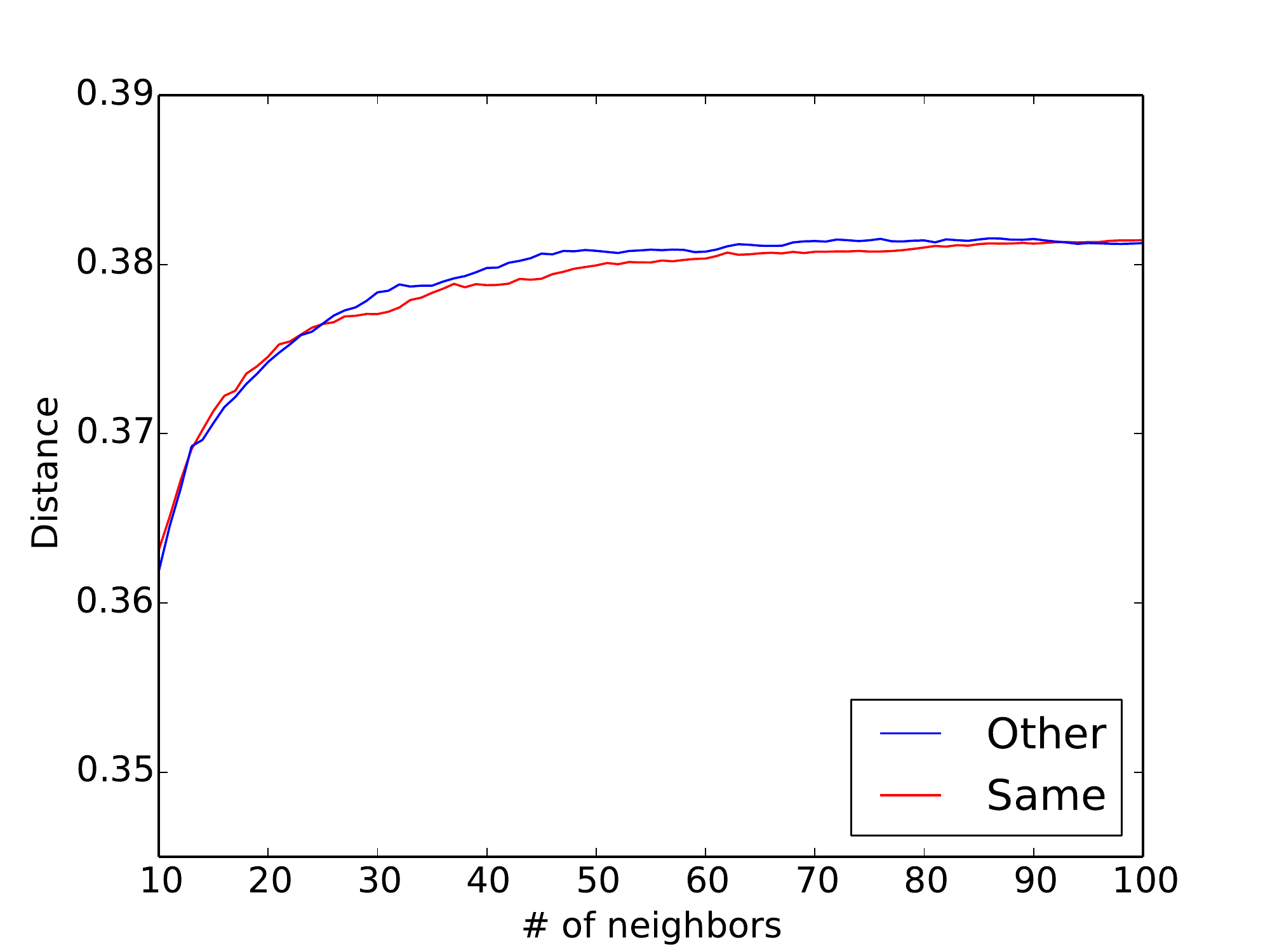}\\
Situation 1
\end{center}
\end{minipage}
\begin{minipage}{0.35\hsize}
\begin{center}
\includegraphics[width=1.0\linewidth]{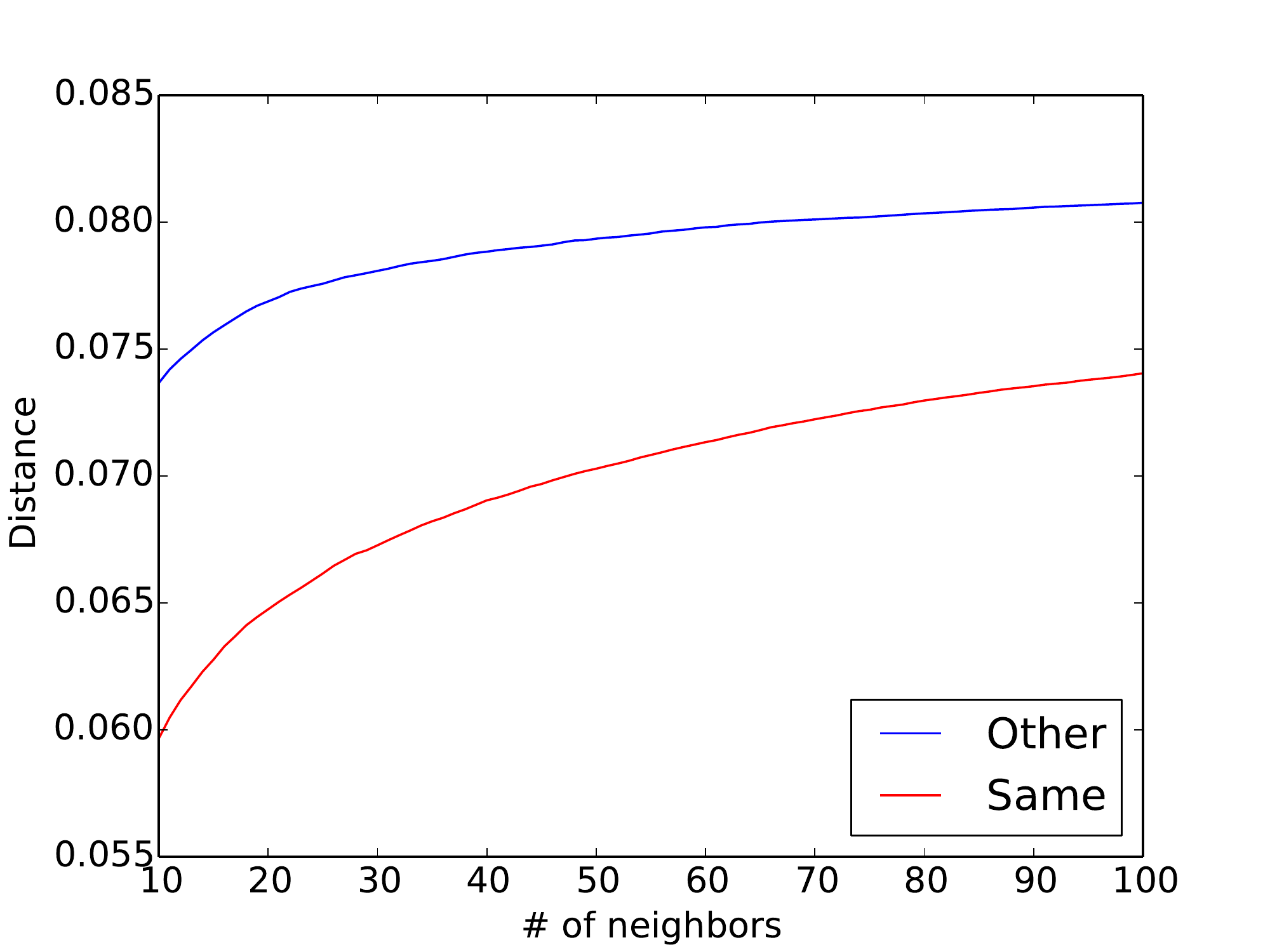}\\
Situation 2
\end{center}
\end{minipage}
\caption{Averaged accumulated Euclidean distance between obstacle's coordinates in $\vect{x}_i$ and $j$-th neighborhood $\vect{x}_i^j$. Red represents $f_u$ of the model with pretraining in the same environment~(Same), blue represents $f_u$ of the model with pretraining in the simple environment~(Other). }
\label{fig:distance_add}
\end{figure}

\begin{figure}[t]
\centering
\begin{minipage}{0.35\hsize}
\begin{center}
\includegraphics[width=1.0\linewidth]{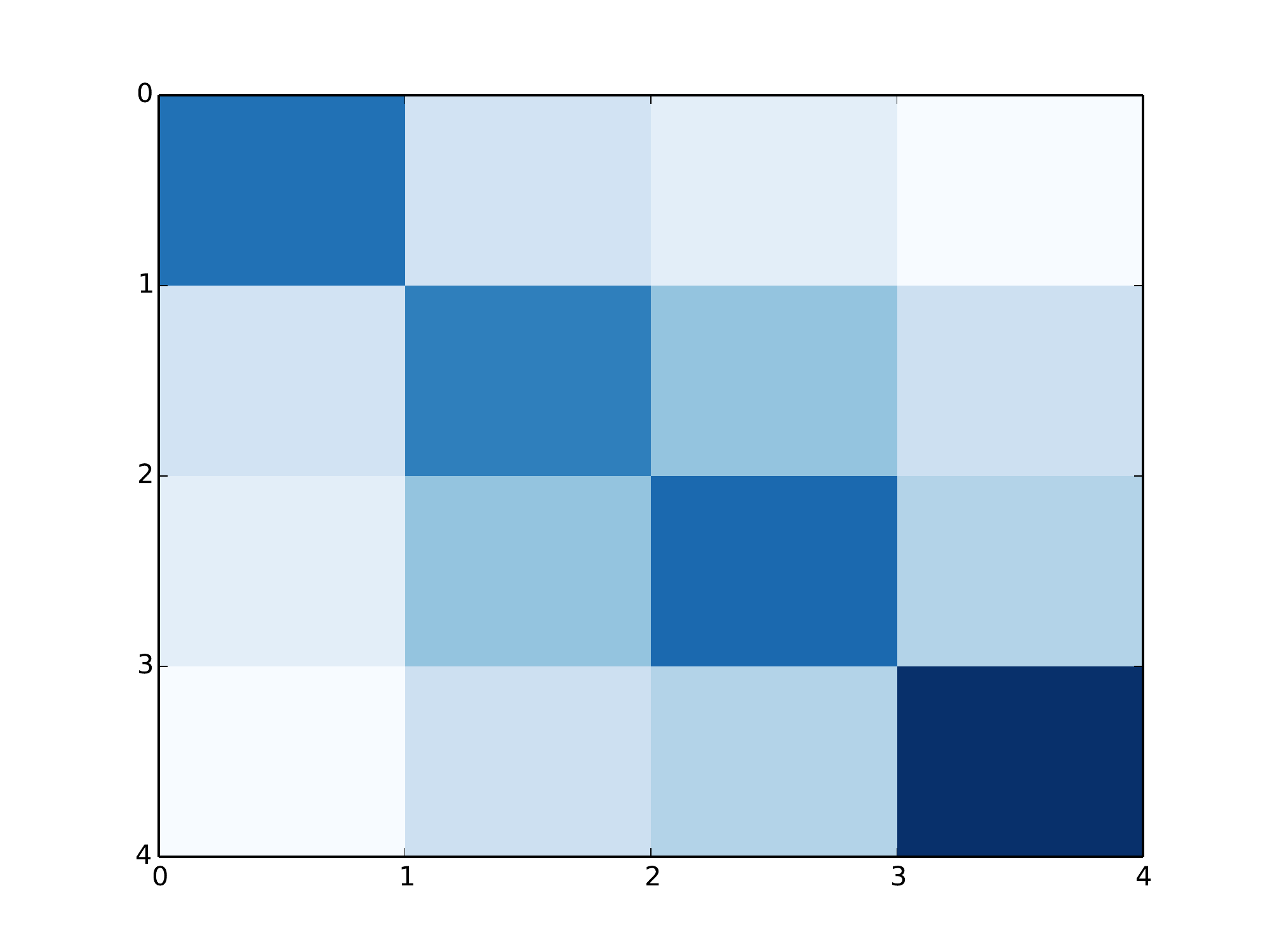}\\
Situation 1
\end{center}
\end{minipage}
\begin{minipage}{0.35\hsize}
\begin{center}
\includegraphics[width=1.0\linewidth]{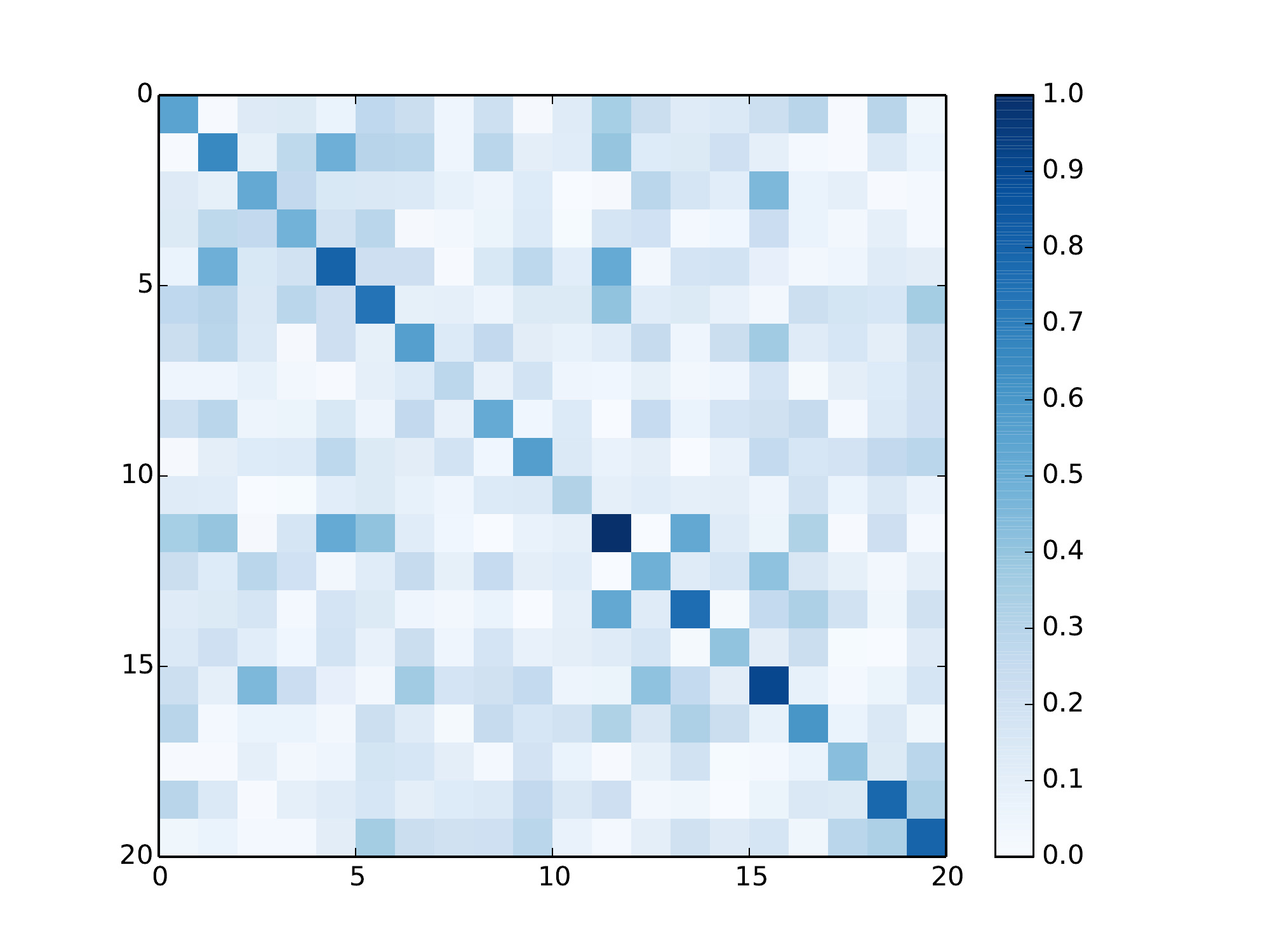}\\
Situation 2
\end{center}
\end{minipage}
\caption{Normalized absolute concentration matrices computed by $f_u(\vect{x})$ using the model with pretraining in the simple environment.}
\label{fig:inverse_add}
\end{figure}

For stable training to disentangle controllable and uncontrollable factors of variation, we could have used other approaches. 
As an example of these approaches, this section present some results obtained when reusing parameters $\vect{\phi}, \vect{\theta}$, and $\vect{\psi}$ from another environment.
We used an environment without uncontrollable obstacles, as shown in Fig.~\ref{fig:env}.

Figures~\ref{fig:weight}, \ref{fig:reconst_reuse}, \ref{fig:reconst_reuse_miss}, \ref{fig:distance_add}, and \ref{fig:inverse_add}, and tables~\ref{tbl:correlation_reusing} and \ref{tbl:correlation_reusing_miss} show results situation 1 and 2.
These results indicate that the model with pretraining in the simple environment has possible to provide similar performance to the model with pretraining in the same environment. 
However, this approach has still room for improvement.
Considering results of situation 2, as an example.
The trained encoder in the simple environment does not have the ability to ignore the obstacles; thus, it does not work well when the environment includes many obstacles.
From this, latent features contain noise and become lower correlation than the model with pretraining in the same environment.
These noisy features increase the reconstruction error and the DNN for obstacles tries to make up for this error.
Therefore, $g_u(f_u(\vect{x}))$ tries to reconstruct the controllable object as part of the uncontrollable obstacles and the distance between obstacle's coordinates in $\vect{x}_i$ and $j$-th neighborhood $\vect{x}_i^j$ increases.
From this point of view, we consider that the curriculum learning~\cite{bengio2009curriculum} which gradually increases the number of uncontrollable obstacles is necessary.



\subsubsection{Results of VAE}

\begin{figure}[t]
\centering
\includegraphics[width=0.65\linewidth]{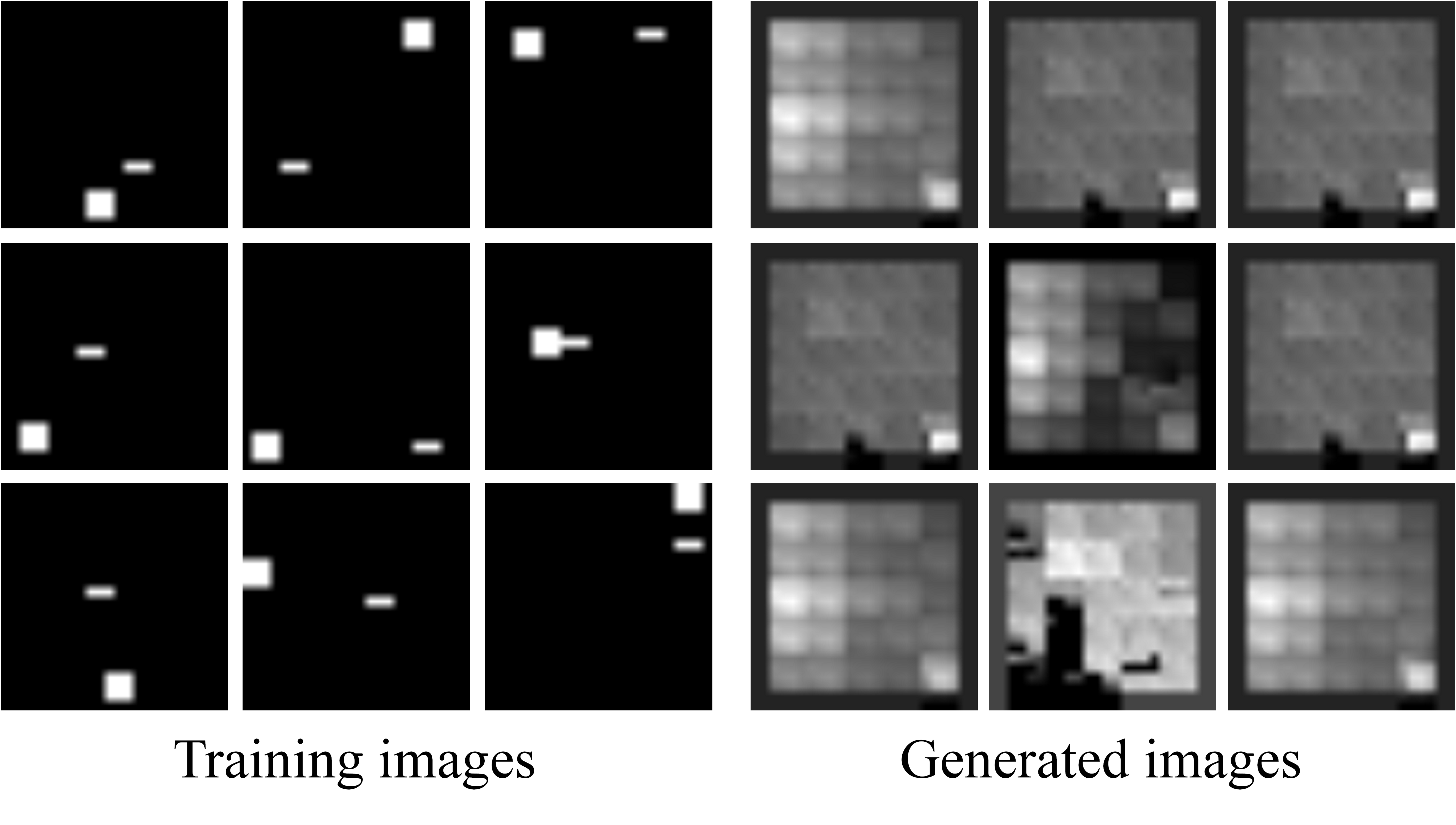}
\caption{Example images generated by the VAE.}
\label{fig:vae}
\end{figure}

\begin{figure}[t]
\centering
\begin{minipage}{0.3\hsize}
\begin{center}
\includegraphics[width=1.0\linewidth]{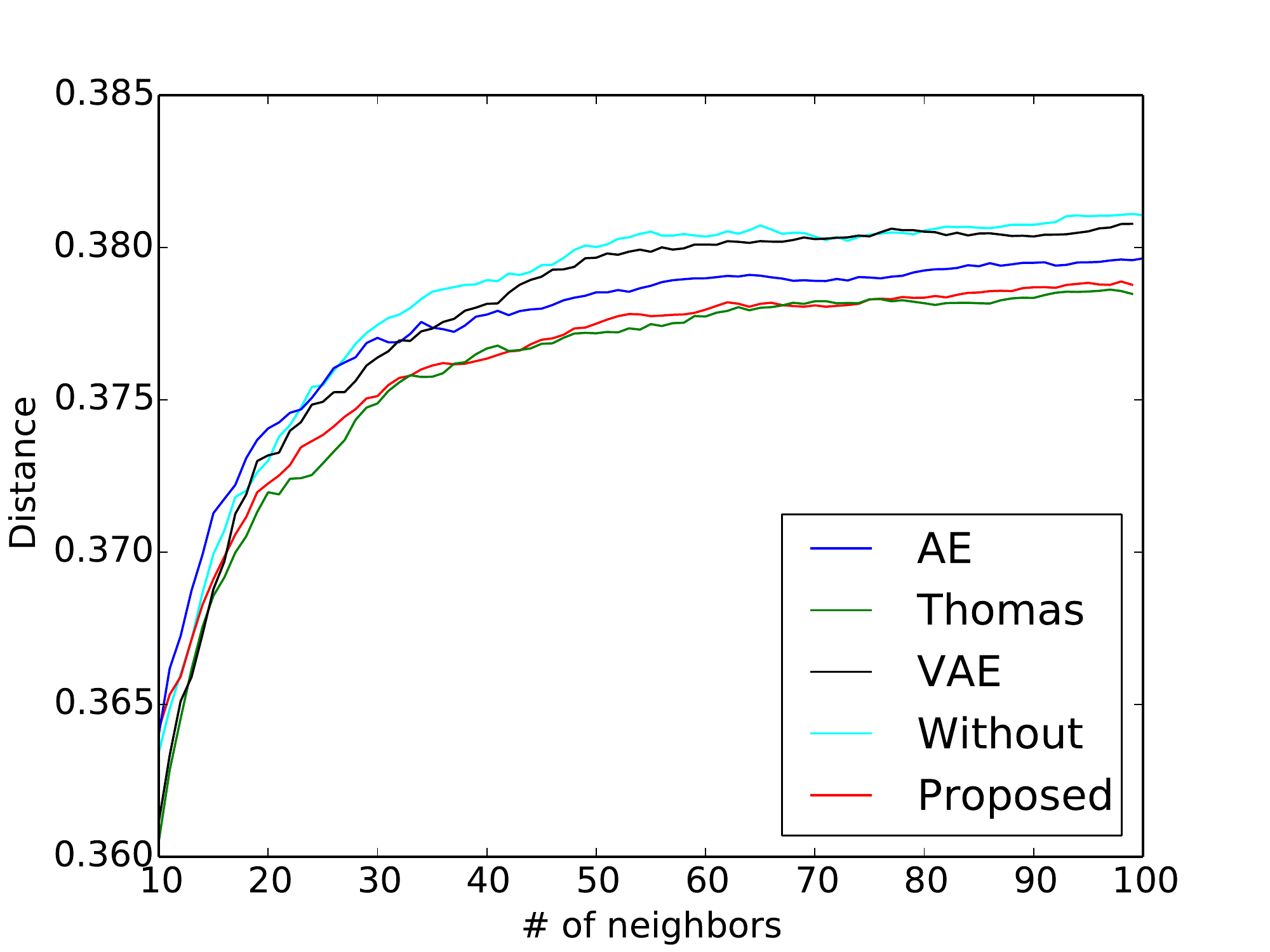}\\
(A)
\end{center}
\end{minipage}
\begin{minipage}{0.3\hsize}
\begin{center}
\includegraphics[width=1.0\linewidth]{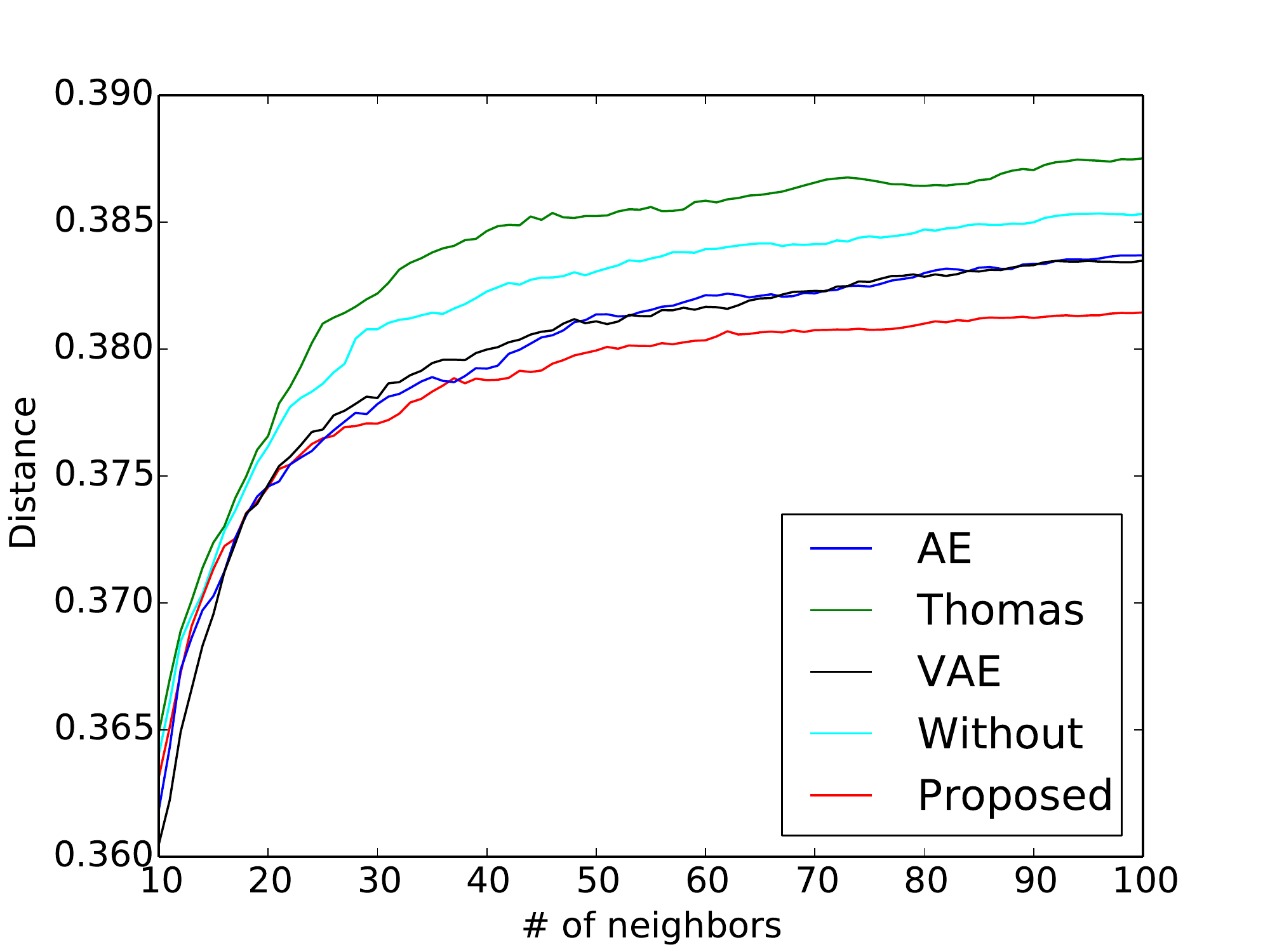}\\
(B)
\end{center}
\end{minipage}
\begin{minipage}{0.3\hsize}
\begin{center}
\includegraphics[width=1.0\linewidth]{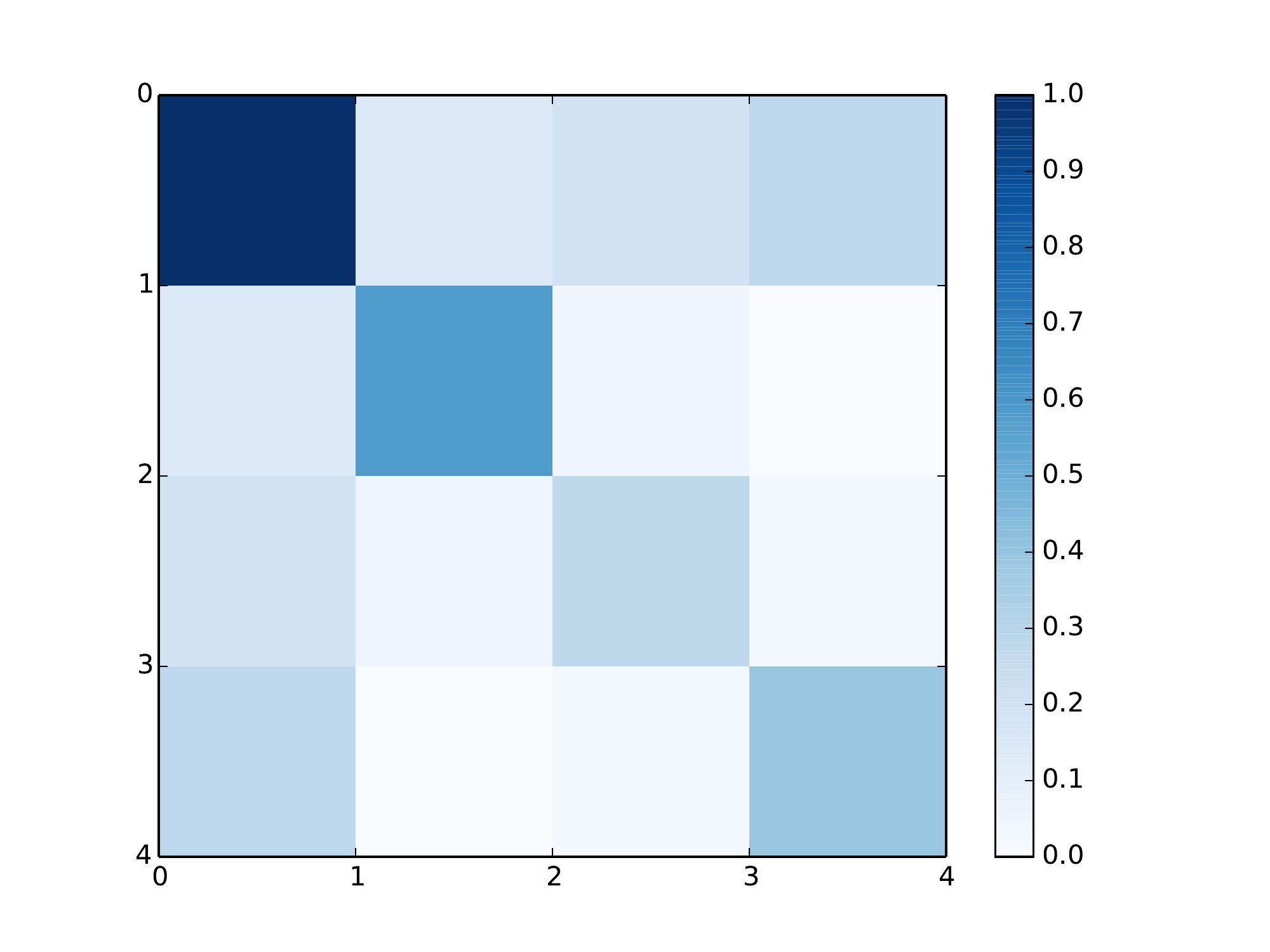}\\
(C)
\end{center}
\end{minipage}
\caption{Averaged accumulated Euclidean distance between (A) object's coordinates and (B) obstacle's coordinates, and (C) normalized absolute concentration matrix computed using the latent features of the VAE.}
\label{fig:vae2}
\end{figure}

Our proposed method is based on the Thomas' method using RL approach to deliver good representations of the environment.
On the other hand, it may be beneficial to show some results of the generation model.
From this point of view, we applied the VAE to situation 1.
Note that the network architecture is the same as the autoencoder.

Figures~\ref{fig:vae} and \ref{fig:vae2} show some results.
As can be observed, the variables of the VAE are independent while the VAE could not generate images well. 
It is known that the VAE tend to produce unrealistic and blurry samples when applied to complex dataset~\cite{chen2016variational,dosovitskiy2016generating,zhao2017towards}.
We consider that the cause is also same for this result.
To solve this problem, we will need deeper analysis and improvement.

{\small

}

\end{document}